\documentclass{article}

\PassOptionsToPackage{numbers, compress}{natbib}

\usepackage[preprint]{neurips_2026}

\usepackage[utf8]{inputenc}
\usepackage[T1]{fontenc}
\usepackage{hyperref}
\usepackage{url}
\usepackage{booktabs}
\usepackage{amsfonts}
\usepackage{amsmath}
\usepackage{amssymb}
\usepackage{nicefrac}
\usepackage{microtype}
\usepackage{xcolor}
\usepackage{graphicx}
\usepackage{subcaption}
\usepackage{siunitx}
\usepackage{multirow}
\usepackage{array}
\usepackage{caption}
\workshoptitle{I Can't Believe It's Not Better (ICBINB): Failure Modes of AI in Biology}

\title{When Quantization Breaks Memory: Recurrent-State Write-Back in Low-Precision Temporal Inference}

\author{%
  Ismail Erbas\thanks{Corresponding author.} \\
  Department of Biomedical Engineering \\
  Center for Modeling, Simulation, and Imaging in Medicine \\
  Rensselaer Polytechnic Institute \\
  Troy, NY 12180 \\
  \texttt{erbasi@rpi.edu} \\
  \And
  Xavier Intes \\
  Department of Biomedical Engineering \\
  Center for Modeling, Simulation, and Imaging in Medicine \\
  Rensselaer Polytechnic Institute \\
  Troy, NY 12180 \\
    \And
  Vikas Pandey \\
  Department of Biomedical Engineering \\
  Center for Modeling, Simulation, and Imaging in Medicine \\
  Rensselaer Polytechnic Institute \\
  Troy, NY 12180 \\
}

\begin{document}

\maketitle

\begin{abstract}
Quantization is widely used to reduce the computational and memory demands of neural-network inference. In recurrent networks, however, the quantized state is stored and returned to the network at the next time step, so the rule used to store that state can alter the subsequent chain of computations. Here, we introduce the term \emph{recurrent-state write-back} to denote this rule and isolate its effect in a compact GRU encoder--decoder for fluorescence lifetime imaging, a molecular imaging modality with applications in quantitative biological imaging. A central task in this setting is estimating two lifetime parameters, the short-lived component $\tau_1$ and the long-lived component $\tau_2$, from high-noise time-resolved fluorescence signals.
Holding the trained model fixed, replacing continuous state propagation with deterministic 4-bit state storage increases the estimation errors for $\tau_1$ and $\tau_2$ by approximately 70-fold and 300-fold, respectively. Failure occurs when repeated small updates remain below the write threshold, leaving the stored state nearly fixed while the network continues to propose change. Error feedback, residual memory, and direction memory carry information from these suppressed updates across time and recover accuracy without retraining.
Precision sweeps show that increasing state precision can worsen a fixed recurrent solution, while matched training shows that compatibility with the state interface can be learned. To test whether this behavior extends beyond the GRU, we repeat the post-training intervention in an independently trained LSTM, where coarse write-back reproduces the failure, error feedback restores accuracy, and state-specific interventions reveal substantially greater sensitivity of the cell state than the hidden state. Our results establish recurrent-state write-back as a key determinant of low-precision recurrent dynamics and identify the state-storage interface as a central design consideration for quantized recurrent inference.
\end{abstract}
\section{Introduction}
\label{sec:introduction}

Quantization has become a standard approach for reducing the memory, arithmetic, and data-movement costs of neural-network inference
\cite{han2015deep,coelho2021automatic,hubara2018quantized}. Its effect is
especially consequential in recurrent networks, where an internal state is
stored after each step and returned to the network to generate the next
\cite{cho2014learning,han2017ese}. The stored state therefore serves as both
memory of the preceding sequence and input to future computation. Changing
how this state is represented can consequently change the sequence of states that the network visits over time. For low-precision recurrent inference, the state-storage operation is therefore part of the temporal computation that is executed at deployment.

Recurrent networks can operate accurately at low precision when the
constraint is present during training: weights, activations, gates, and
recurrent states can all be learned at reduced bit widths
\cite{xu2018alternating,hou2019normalization}. Because the numerical
representation is fixed before learning begins, the network forms its
dynamics around it, and quantization-aware training is therefore often
treated as the remedy for low-precision degradation. Two observations show
that the remedy is incomplete. First, deployment can reverse the order of
constraint and learning: a trained network may be mapped to a coarser
representation to satisfy computational or hardware constraints, without
retraining \cite{jacob2018quantization,salah2025post}. Every stored state
is then written through a numerical interface the network never trained
with, and a storage rule its learned dynamics do not expect can withhold
the state updates they depend on. Second, even when the constraint is
present throughout training, coarse state precision can leave an accuracy
gap that further optimization does not close, and the mechanism behind the
gap has not been identified. Both observations point to a single operation:
the rule that stores the recurrent state between steps. The question is
then direct: what does the state-storage rule do to the temporal behavior
of a trained recurrent network, and can it produce failures that bit width
alone does not explain?

We call the rule that determines this stored value \emph{recurrent-state write-back}: the mapping from the state computed at recurrent step $t$ to the value stored and presented to step $t+1$.
Deterministic low-precision write-back rounds the computed state to the nearest representable level. Updates that remain within the corresponding write boundary leave the stored state unchanged and therefore do not enter the state observed by the next recurrent step. This becomes consequential when the network repeatedly proposes small changes in the same direction across successive steps. The computed recurrence can continue to propose movement while the stored trajectory remains unchanged. Recurrent-state write-back therefore determines when computed state changes become visible to future computation and how temporally distributed information is carried through a quantized recurrence.

We study this problem in fluorescence lifetime imaging (FLI), where quantitative molecular information is encoded in the temporal profile of fluorescence emission
\cite{lakowicz2006principles,becker2012fluorescence,suhling2015fluorescence,smith2023vivo}.
FLI provides sensitivity to molecular interactions and local
microenvironmental changes, with demonstrated applications in tumor visualization, targeted drug delivery, and molecular target engagement \cite{unger2020real,yuan2024antibody,verma2024fluorescence,rudkouskaya2020multiplexed,smith2023vivo}.
In biomedical imaging settings that require rapid feedback, including image-guided intervention, time-resolved fluorescence measurements must be
converted efficiently into quantitative lifetime estimates, motivating
compact models capable of low-latency inference. Each pixel provides a
high-noise time-resolved fluorescence signal whose temporal structure
reflects fluorescence decay kinetics, photon statistics, and the acquisition
response. The inference targets are the short- and long-fluorescence
lifetimes, $\tau_1$ and $\tau_2$; for sample $i$, we denote the corresponding
parameter vector by
$\boldsymbol{\theta}_i=(\tau_{1,i},\tau_{2,i})$.

The recurrent model used here is Seq2SeqLite, the compact student
architecture derived from the Seq2Seq framework for time-resolved
fluorescence reconstruction and lifetime estimation
\cite{pandey2024deep,erbas2024compressing,erbas2024unlocking}. Seq2SeqLite
uses a single-layer 32-unit GRU encoder--decoder with 6,627 trainable
parameters and was selected in the associated compression and deployment
studies for low-precision inference. Those studies document the second
observation above: 8-bit precision was adopted as the operating point
because 4-bit inference fell short of the required accuracy even when the
constraint was present during training, and the mechanism behind the
shortfall was not identified. Seq2SeqLite therefore provides a natural
setting for the present question: its compact recurrent state and
documented 4-bit shortfall allow the state-storage interface to be examined
while retaining the same fluorescence-lifetime inference task. The encoder
processes 135 temporal samples and compresses the measured signal into a
recurrent state; the decoder reconstructs deconvolved temporal output
sequences whose integrated shape yields the lifetime estimates
$\hat{\boldsymbol{\theta}}_i=
(\hat{\tau}_{1,i},\hat{\tau}_{2,i})$. Because these parameter
estimates depend on temporal structure accumulated across the sequence, FLI example
provides a sensitive setting for examining how recurrent-state storage
affects temporally distributed information.

This setting lets us test, under controlled conditions, whether the
state-storage rule can alter a trained recurrent network by withholding
proposed state changes from the state visible to future recurrent steps.
To isolate storage from learning, we intervene on trained networks without
retraining them: each checkpoint is reconstructed, verified to reproduce
its native implementation, and re-evaluated with only the recurrent-state
write-back rule replaced, while the trained weights, recurrent update,
readout, inputs, and evaluation procedure remain fixed. If storage alone is
responsible, this single change should break the network, the failing
trajectory should show the predicted suppression of proposed state changes,
and preserving the suppressed information should restore accuracy in the
same fixed network; all three predictions are examined directly. We then
separate state precision from state-interface compatibility, by changing
precision after training and by training matched models around different
recurrent-memory interfaces, and finally repeat the post-training
intervention in an independently trained 32-unit LSTM to ask whether the
mechanism extends beyond the GRU and whether its consequences depend on
which recurrent variable stores the affected information.

This study establishes recurrent-state write-back as a distinct design
variable in low-precision recurrent inference. By changing only the storage
rule while holding the trained computation fixed, we isolate a direct causal
link between state write-back and the trajectory a recurrent network
executes, and we identify persistent suppression of proposed state changes
as the temporal mechanism behind the resulting degradation. Two tools make
the mechanism measurable and correctable. The \emph{recurrent write
margin} compares each proposed recurrent-state change with the half-step of
the state quantizer and so indicates directly whether that change can enter
the recurrence-visible trajectory under deterministic write-back.
\emph{Direction memory}, introduced here, retains the direction of
persistent sub-threshold proposed changes in a compact auxiliary counter
and converts accumulated same-direction evidence into a later state
transition; together with error feedback and residual memory, it restores
accurate inference in fixed networks without retraining. The central result
is that write-back is part of the temporal computation a deployed network
executes, not a passive encoding of a trajectory the network would produce
anyway: precision sweeps and matched training show that the relevant
property is not bit width alone but the compatibility between a learned
recurrent solution and the interface through which it is executed, and an
independently trained LSTM reproduces the failure, its rescue, and a strong
dependence on which recurrent variable stores the affected information.
\section{Methods}
\label{sec:methods}
We describe the inference task and recurrent models, then the write-back
interventions, memory operators, precision sweeps, matched training, and
LSTM analysis.

\begin{figure}[t]
\centering
\includegraphics[width=\textwidth]{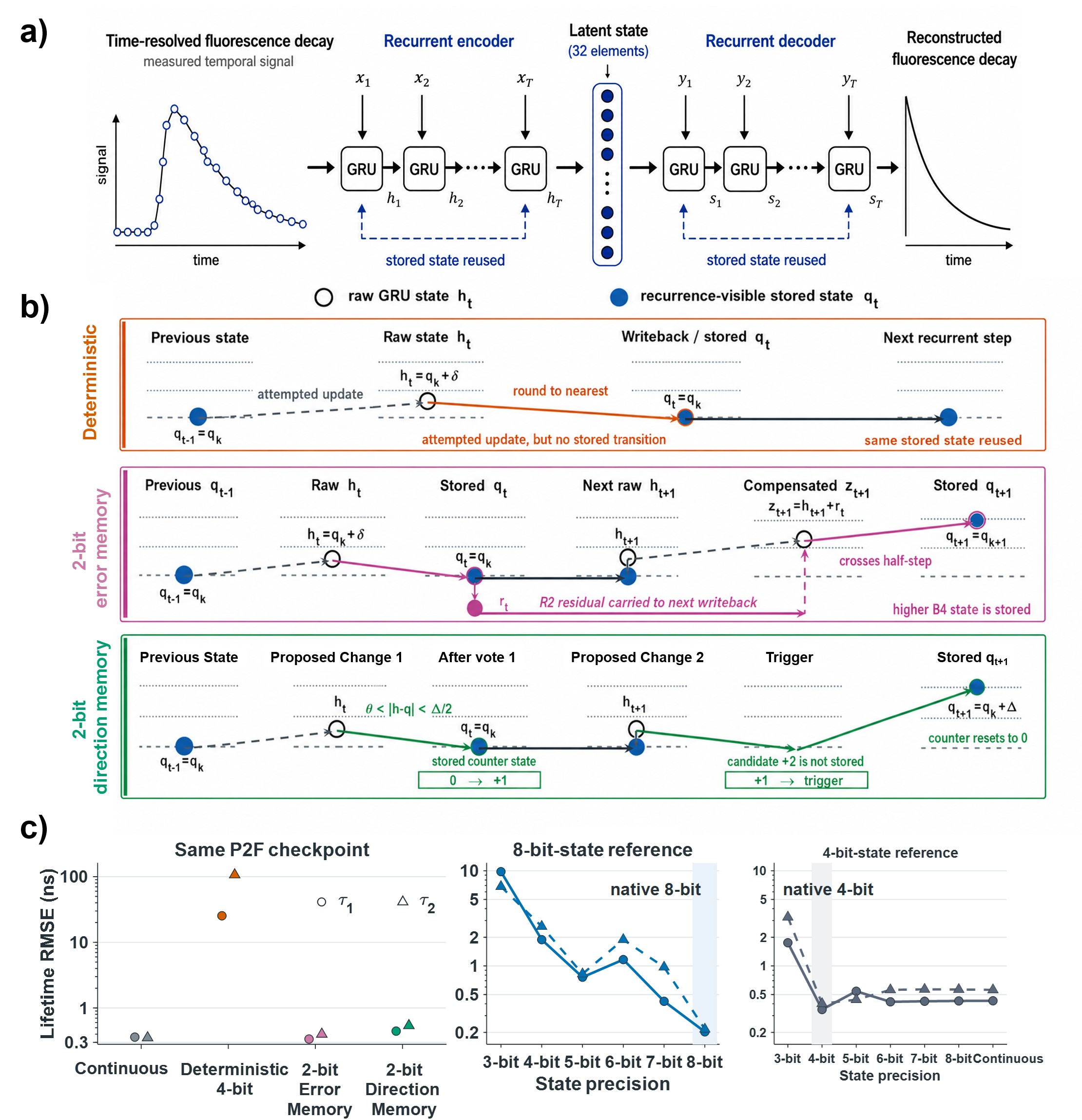}
\caption{\textbf{Recurrent-state write-back in low-precision temporal
inference.}
\textbf{a,} Seq2SeqLite: a 32-unit GRU encoder processes a high-noise
time-resolved fluorescence signal and initializes a 32-unit recurrent
decoder; the state is stored after every recurrent step and returned to the
next.
\textbf{b,} Deterministic 4-bit write-back rounds each computed state to the
nearest representable level, so proposed state changes below the half-step
write boundary leave the stored state unchanged. Residual memory carries the
discarded magnitude across steps; direction memory accumulates repeated
same-direction proposals.
\textbf{c,} Lifetime RMSE for post-training write-back interventions on the
QMem P2F checkpoint (trained with 4-bit state write-back) and state-precision
sweeps of independently trained 4-bit-state and 8-bit-state reference GRUs.
Circles: short lifetime $\tau_1$; triangles: long lifetime $\tau_2$; RMSE in
nanoseconds, logarithmic axes.}
\label{fig:overview}
\end{figure}

\subsection{Fluorescence lifetime inference and recurrent models}
\label{sec:methods_fli}

The study uses the simulated fluorescence-lifetime dataset associated with the Seq2SeqLite model family \cite{erbas2024compressing,pandeypyfli}. The dataset
contains 1,600,000 high-noise time-resolved fluorescence signals, each represented by 135 temporal bins.
A fixed 80/10/10 partition provides 1,280,000 decay samples for training, 160,000 for validation, and 160,000 held out for testing. The same partition is used throughout the GRU training, post-training interventions, precision analyses, and matched-training comparisons.

Seq2SeqLite is a single-layer 32-unit GRU encoder--decoder with a linear
readout and 6,627 trainable parameters. The encoder processes the 135-step fluorescence decay signal and passes its final recurrent state to the
decoder, which generates the temporal output sequence. The 32-unit model is trained by knowledge distillation from a frozen 128-unit teacher using the established Seq2SeqLite framework \cite{erbas2024compressing}.  Lifetime estimates are calculated from the predicted temporal outputs by
trapezoidal integration normalized to the first-gate amplitude and compared
with the ground-truth $\tau_1$ and $\tau_2$ labels using RMSE (Supplementary Methods~\ref{sec:supp_lifetime}). Sequence MAE reports the mean absolute difference between the predicted and target
temporal sequences and is used to distinguish reconstruction error from error in the derived lifetime parameters. Full data
generation, normalization, model-training, and lifetime-extraction procedures are provided in Supplementary Methods~\ref{sec:supp_training} and \ref{sec:supp_lifetime}.

We developed QMem, a staged quantization-aware training (QAT) procedure,
to obtain checkpoints immediately before, during, and after the
recurrent-state interface is hardened (\autoref{fig:supp_qmem}). QMem
introduces 4-bit quantization sequentially for the candidate kernels (P2A),
reset-gate kernels (P2B), update-gate kernels (P2C), biases (P2D),
candidate activation (P2E), and finally the recurrent state (P2F), with
each quantizer retained once introduced. During P2F, the
recurrence-visible state $q_t$, the stored state received by the next
recurrent step, moves progressively onto the 4-bit grid over the first
15 epochs, after which full deterministic 4-bit write-back is retained; P3
then continues training on the fully quantized inference graph. In the
analysis, P2E is the network immediately before state quantization, P2F is
the primary checkpoint for testing whether changing write-back alone
produces failure, and P3 shows how further optimization adapts to the hard
interface. QMem is developed as a diagnostic scaffold, not the object of the mechanistic claim.

Two independently trained reference GRUs separate this checkpoint-specific
analysis from the broader question of state-interface compatibility. An
8-bit-state reference model tests whether a network trained around a finer
state interface can be disrupted by post-training 4-bit write-back and rescued
without changing its weights. A 4-bit-state reference model provides the
complementary control: it tests whether training around the 4-bit interface
avoids the write-back failure, and whether
increasing state precision after training necessarily improves the same fixed
solution. Full training, quantizer, and checkpoint-selection details are given
in Supplementary Methods~\ref{sec:supp_training}.

\subsection{Post-training recurrent-state write-back intervention}
\label{sec:methods_frozen}

To isolate the recurrent-state storage operation from learning, each trained
checkpoint is held fixed while only the rule that stores the recurrent state
between successive time steps is changed. Learned weights and biases,
recurrent gate and candidate-state calculations, dense readout, input
signals, and the evaluation procedure remain unchanged, and no retraining or
fine-tuning is performed. We refer to this controlled manipulation as a
\emph{post-training recurrent-state write-back intervention}. A native
condition retains the state-storage rule used by the trained model, whereas
a post-training condition changes only this interface to an alternative
write-back rule, such as deterministic 4-bit storage, finer state precision,
or continuous propagation.

The recurrence-visible state $q_{t-1}$ enters recurrent step $t$, the GRU
computes the raw state $h_t$ before storage, and a deterministic $B$-bit
write-back operator with grid spacing $\Delta_B$ stores $q_t=Q_B(h_t)$. For hidden unit $j$, the proposed
state change is
$\delta_{t,j}=h_{t,j}-q_{t-1,j}$. To quantify the relation between the GRU state update and the quantizer write
boundary, we introduce the \emph{recurrent write margin},
\begin{equation}
M_{t,j}
=
\frac{2|\delta_{t,j}|}{\Delta_B}
=
\frac{
2(1-z_{t,j})
|\tilde{h}_{t,j}-q_{t-1,j}|
}{
\Delta_B
},
\label{eq:write_margin}
\end{equation}
where $z_{t,j}$ is the GRU update gate and $\tilde{h}_{t,j}$ is the
candidate state. For an interior grid level away from a rounding tie,
$M_{t,j}<1$ places the proposed state change inside the half-step write
boundary, so deterministic nearest-level write-back retains the same stored
value. We refer to this region as the \emph{write deadband} and define the
deadband fraction as the fraction of proposed state changes satisfying
$M_{t,j}<1$. The complete GRU recurrence, write-back
operators, rail and tie handling, and diagnostic definitions are provided in
Supplementary Methods~\ref{sec:supp_write-back}.
\subsection{Temporal-memory and state-precision interventions}
\label{sec:methods_memory}

Post-training write-back interventions differ in how information is
retained between recurrent steps while the trained network remains fixed.
Continuous propagation returns the computed state directly; stochastic
rounding selects between adjacent representable levels; error feedback
carries the discarded quantization error into the next write opportunity;
quantized residual memory stores the discarded magnitude in a $k$-bit
auxiliary state; and direction memory, introduced here, accumulates the
sign of repeated sub-threshold proposed changes in a $k$-bit counter and
advances the stored state by one quantization level when the accumulated
same-direction evidence reaches the counter threshold. Throughout, $B$
denotes the bit width of the recurrence-visible state and $k$ that of an
auxiliary memory, so the operators can be compared while the state
presented to the recurrent network remains 4-bit. Complete operator
definitions and auxiliary-state handling are given in Supplementary
Methods~\ref{sec:supp_write-back}.

State precision is examined by evaluating independently trained 4-bit and 8-bit GRUs under alternative post-training state-write-back precisions while keeping the trained parameters, recurrent computation, readout, inputs, and evaluation fixed. The 4-bit-state reference GRU is also evaluated with continuous state propagation, and per-unit state occupancy is measured to determine whether changes in numerical resolution are expressed in the recurrent trajectory. Two further analyses test whether interface compatibility can be learned
and whether the mechanism extends beyond the GRU. In matched training,
four recurrent-memory configurations are trained from identical
initializations, a 4-bit state, a 6-bit state, a 4-bit state with 2-bit
residual memory, and a 4-bit state with 2-bit direction memory, with input
kernels, recurrent kernels, biases, and candidate activations held at
4-bit throughout, so only the allocation of recurrent-state memory
changes. In the cross-architecture analysis, the post-training
intervention is repeated in an independently trained 32-unit LSTM with
native 8-bit cell- and hidden-state write-back, changing write-back for
both recurrent variables together or each individually. Complete procedures are in
Supplementary
Methods~\ref{sec:supp_trained_campaign}.

\section{Results}
\label{sec:results}

The results follow the study's causal chain: establish that changing
write-back alone breaks a fixed trained network, identify the temporal
signature of that failure, rescue the same frozen network by preserving
the suppressed information, then test whether bit width, learned
compatibility, or architecture governs the effect. Throughout, paired errors are reported as $\tau_1/\tau_2$ lifetime RMSE in nanoseconds.

\subsection{Post-training recurrent-state write-back intervention}

The progressive QMem trajectory first identifies the point at which recurrent
state becomes sensitive to coarse storage. P2E retains continuous recurrent
state and reaches $\tau_1/\tau_2$ RMSE of 1.40/3.03~ns. Introducing 4-bit
recurrent-state write-back at P2F increases these errors to 25.37/106.59~ns,
after which P3 fine-tuning recovers 0.48/0.55~ns (Supplementary
Fig.~\ref{fig:supp_phase_predictions}). Because the network continues to train
between these phases, this transition shows where the failure appears but does
not by itself establish that state write-back is the cause.

We therefore returned to the P2F checkpoint and changed only the value stored
between recurrent steps. With the checkpoint fixed, continuous identity
propagation gives 0.36/0.35~ns lifetime RMSE. Replacing only this state-storage
rule with deterministic 4-bit write-back gives 25.37/106.59~ns
(Fig.~\ref{fig:overview}c). For this checkpoint, the roles are reversed
relative to the later experiments: 4-bit write-back is the native interface
introduced at P2F, so identity propagation is the intervention; either
direction of the comparison isolates the same single operation. The trained parameters, recurrent update,
readout, input signals, and evaluation procedure are identical in the two
conditions. The large change in task performance therefore arises from what is
stored and returned to the next recurrent step. The failure is not equally apparent in the reconstructed sequence itself:
deterministic P2F retains a sequence MAE of 0.09 while the lifetime
estimate is severely inaccurate, so a small pointwise sequence error can
distort the temporal shape from which lifetime is calculated
(Supplementary Results~\ref{sec:supp_metric_interpretation}).

\subsection{Recurrent-state write activity and temporal organization}

Having isolated write-back as the changing operation, we next examined how the
stored state evolves during the failing computation. At the fixed P2F checkpoint, 99.59\%
of decoder updates lie inside the 4-bit write boundary. Only 0.25\% of decoder
state elements change stored level between successive steps, 97.25\% of
decoder steps change no hidden unit at all, and an average of only 0.08 of the
32 decoder units changes level at each step (Supplementary Tables~\ref{tab:write-back_summary} and \ref{tab:margin_summary}; Supplementary Fig.~\ref{fig:supp_per_unit}). The recurrent network is therefore
continuing to compute new state proposals, but most of those proposals do not
become part of the state that is carried forward.

\begin{figure*}[t]
\centering
\includegraphics[width=\textwidth]{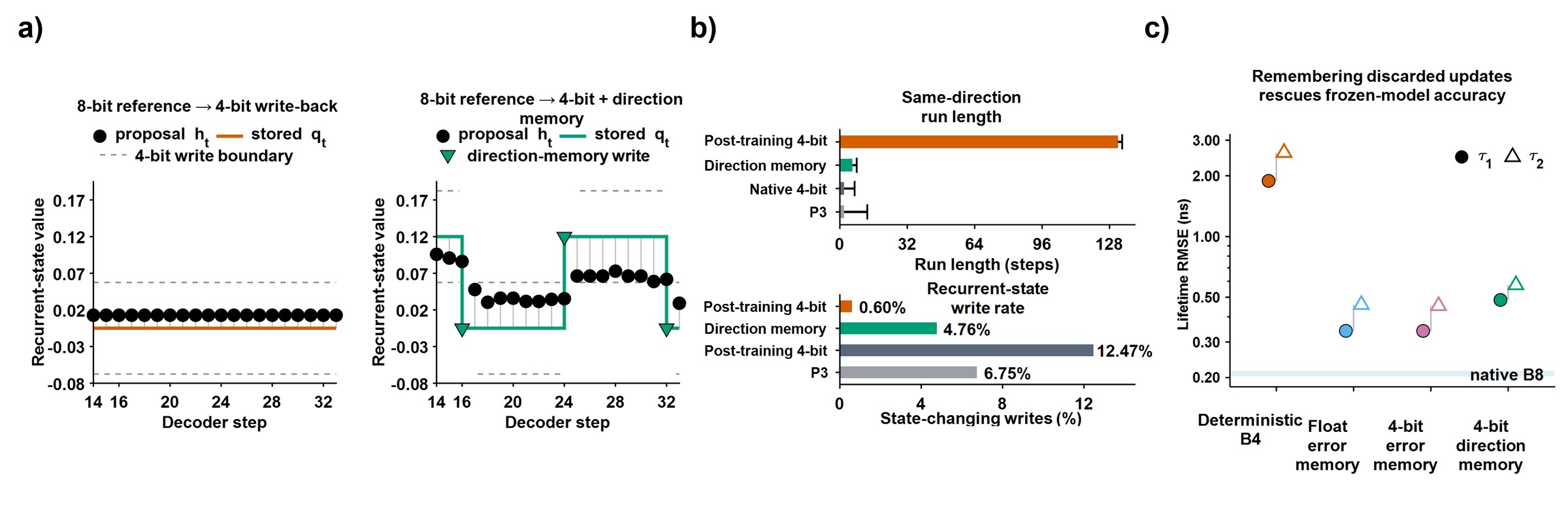}
\caption{\textbf{Temporal memory restores recurrent-state changes suppressed
by coarse write-back.}
\textbf{a,} Decoder-state trajectories from the same 8-bit-state reference
GRU under post-training deterministic 4-bit write-back and 4-bit write-back
with direction memory. Circles: raw computed state; colored lines: stored
state returned to the next step; dashed lines: 4-bit write boundaries;
triangles: direction-memory state transitions.
\textbf{b,} Same-direction runs of consecutive sub-threshold proposed state
changes (bars: median completed run length; whiskers: to the 90th
percentile) and the mean fraction of decoder state elements changing per
step. The 4-bit-state reference GRU and QMem P3 provide trained-interface
comparisons.
\textbf{c,} Lifetime RMSE for the 8-bit-state reference GRU under native
8-bit write-back and post-training 4-bit write-back with deterministic
rounding, error feedback, residual memory, or direction memory. Circles:
$\tau_1$; triangles: $\tau_2$; shading spans the native 8-bit RMSE. Panels
\textbf{b} and \textbf{c} cover the complete 160,000-sample held-out test
partition.}
\label{fig:write-back_mechanism}
\end{figure*}

The independently trained 8-bit-state reference GRU provides a direct
post-training comparison because the same checkpoint can be evaluated with
its native 8-bit state interface or with deterministic 4-bit write-back.
Changing only recurrent-state write-back from the native 8-bit interface to
deterministic 4-bit reduces the decoder state-change fraction to 0.60\%. The suppressed updates
are highly organized in time: consecutive sub-threshold updates almost always
point in the same direction, with same-sign fractions above 0.9989 across the
GRU conditions examined (Supplementary Table~\ref{tab:sign_persistence}). Directional persistence alone, however, is not a
marker of failure because it is also present in accurate models. The important
difference is how long those updates remain absent from the stored trajectory.
Under post-training 4-bit write-back, the median completed same-direction run lasts 132
of the 134 live decoder write steps and the 90th percentile reaches 134 steps.
In contrast, the corresponding median is 2 steps for both native 4-bit and P3,
where recurrence-visible state changes repeatedly interrupt these runs
(Fig.~\ref{fig:write-back_mechanism}b). Native 4-bit changes 12.47\% of decoder
state elements per step despite having many sub-threshold updates. These
comparisons show that coarse quantization does not fail simply because small
updates exist. It fails when the storage rule prevents persistent proposed
changes from entering the recurrent trajectory for a substantial fraction of
the computation.

\subsection{Temporal-memory write-back interventions}

If the failure is caused by useful state updates being repeatedly discarded,
then preserving information from those updates should improve the same fixed
network without changing its weights. This is what we observe. At P2F, error
feedback reduces lifetime RMSE to 0.36/0.37~ns, a 2-bit residual memory reaches
0.34/0.40~ns, and 3-bit direction memory reaches 0.34/0.34~ns (Supplementary Table~\ref{tab:temporal_memory_sweep}). These
operators retain different information between write opportunities. Error
feedback and residual memory preserve update magnitude, whereas direction memory preserves accumulated direction. In all cases, the state returned to the
recurrent network remains on the 4-bit grid. Their common effect is therefore not
higher state precision, but the ability to carry information that cannot be
written at one step forward to a later write opportunity.

The same pattern appears in the independently trained 8-bit reference. Native 8-bit
reaches 0.20/0.22~ns. Forcing only its recurrent state onto deterministic 4-bit
increases the error to 1.89/2.60~ns and raises the decoder deadband fraction to
99.40\%. On that same fixed checkpoint, error feedback recovers 0.34/0.46~ns,
4-bit residual memory reaches 0.34/0.45~ns, and 4-bit direction memory reaches
0.49/0.58~ns (Fig.~\ref{fig:write-back_mechanism}c). Direction memory also
shortens the median completed same-direction run from 132 to 6 steps and raises
the decoder state-change fraction from 0.60\% to 4.76\%
(Fig.~\ref{fig:write-back_mechanism}a,b). The ability of distinct memory
mechanisms to rescue the same fixed failure shows that the discarded updates
are not merely numerical noise. Information about those updates remains useful
to later recurrent computation.

This rescue has a boundary. The independently trained native-4-bit model
is already accurate under deterministic 4-bit write-back (0.35/0.40~ns),
and none of the memory or continuous-state alternatives improves both
lifetime parameters over its native interface. Temporal memory is
therefore not intrinsically better than deterministic storage; it helps
when the storage rule removes information the recurrent solution depends
on.

\subsection{Recurrent-state precision and interface adaptation}
The rescue experiments raise a broader question: if coarse state storage can
harm one trained model, should increasing state precision always help? The
post-training precision sweeps show that it does not, in either direction.
The 8-bit-state reference GRU performs best near its native 8-bit interface
and degrades under post-training 4-bit write-back. The 4-bit-state reference
GRU shows the complementary behavior: it avoids the write-back failure but
remains short of the native 8-bit reference, and increasing only its state
precision to 8-bit does not close this gap. Instead, performance worsens
from 0.35/0.40~ns to 0.43/0.57~ns, with continuous propagation at
0.43/0.56~ns (Fig.~\ref{fig:state_precision_alignment}a). The finer
representation is not simply ignored: across the 32 decoder units, the
median number of occupied state levels increases from 12 under 4-bit to
176.5 under 8-bit (Supplementary Table~\ref{tab:per_unit_summary}), yet the
lifetime estimate becomes less accurate
(Fig.~\ref{fig:state_precision_alignment}b). \emph{Numerical fidelity}, how
finely an individual state value is represented, is therefore distinct from
\emph{dynamical fidelity}, whether repeatedly writing and returning those
stored values produces a trajectory compatible with the learned solution,
and increasing the first can reduce the second for a fixed recurrent
solution.

Matched training provides the complementary test. Four recurrent-memory
allocations were trained from matched initializations under a common
specification: a 4-bit recurrence-visible state, a 6-bit recurrence-visible
state, a 4-bit state with 2-bit residual memory, and a 4-bit state with
2-bit direction memory. When the interface is present throughout
optimization, the post-training ranking changes: direction memory gives the
lowest mean $\tau_1$ RMSE and improves on the 4-bit-state and 6-bit-state
conditions in all three matched runs, whereas the 6-bit-state and
residual-memory conditions remain stronger on $\tau_2$
(Fig.~\ref{fig:state_precision_alignment}c; Supplementary
Tables~\ref{tab:trained_campaign} and \ref{tab:trained_campaign_paired}).
The recurrent solution and its state interface can therefore adapt to one
another during training; the campaign establishes learned compatibility
rather than a universal ranking of memory operators.
\subsection{Cross-architecture LSTM replication and state-specific analysis}

The GRU results suggest that write-back sensitivity is a property of
recurrent computation rather than of one quantized checkpoint. We
therefore repeated the post-training intervention in an independently
trained 32-unit LSTM, whose two recurrent variables play different roles:
the cell state $c$ and the hidden state $h$. Under its native
deterministic 8-bit interface the LSTM reaches 0.239/0.254~ns; forcing
both states to deterministic 4-bit raises this to 3.857/1.327~ns, 4-bit
error feedback on both states restores 0.263/0.332~ns, and continuous
propagation gives 3.501/1.749~ns rather than recovering the native
solution (Fig.~\ref{fig:lstm_replication}a). As in the GRU, a numerically
finer interface is not by itself sufficient to recover the learned
behavior.

The two states are not equally sensitive. Forcing only $c$ to 4-bit (with
$h$ at 8-bit) raises RMSE to 8.158/1.292~ns, whereas forcing only $h$
gives 0.296/0.354~ns (Fig.~\ref{fig:lstm_replication}b). Deadband
frequency alone cannot explain this: the $c$-only intervention places
79.43\% of decoder cell-state updates inside the write boundary with a
90th-percentile same-direction run of 121 of 134 steps, and targeted error
feedback shortens this run to 4 steps and restores 0.257/0.280~ns, while
the $h$-only intervention has a larger deadband fraction of 93.78\% yet a
much smaller task error. The consequence of suppressed updates therefore
depends on which recurrent variable carries the information, not only on
how often or how long suppression occurs.

\section{Discussion}

Quantization is usually evaluated by asking how much numerical precision
can be removed before a model loses accuracy. For recurrent networks that
framing is incomplete: the quantized state is repeatedly returned to the
model and becomes part of the computation that generates its future
states. At the P2F checkpoint, changing only the state-storage rule moves
lifetime RMSE from 0.36/0.35~ns to 25.37/106.59~ns while every trained
parameter and all other operations are unchanged. Recurrent-state storage
is therefore not a passive encoding of a trajectory the network would
otherwise produce; it helps determine the trajectory itself.

The state-level analysis explains how this failure develops over time.
Coarse write-back is not harmful simply because many individual updates
are smaller than one quantization step. The native-4-bit model, which avoids
this failure, also contains many such updates. It becomes harmful when persistent
same-direction proposals never reach the recurrence: in the failing
condition, the stored state remains unchanged for almost the complete
decoder sequence, so the network repeatedly receives a state from which a
persistent component of its required temporal evolution has been removed.

The temporal-memory interventions test this interpretation directly.
Error feedback, quantized residual memory, and direction memory retain
different aspects of the discarded information, yet all recover
substantial accuracy in a frozen network whose 4-bit write-back fails.
Their shared feature is persistence across write opportunities:
information that cannot affect the stored state immediately is allowed to
influence a later transition, which is why even direction memory, which
retains no magnitude, recovers much of the lost performance. The relevant
design question is therefore not only how closely each stored value
approximates the computed state, but how information that cannot be
represented immediately is handled across time.

The precision sweeps and matched training sharpen the distinction between
numerical and dynamical fidelity introduced above. A finer state
interface improves pointwise resolution yet can produce a trajectory less
compatible with the learned solution, while an interface present
throughout optimization lets the recurrent solution adapt to the
transitions that interface allows. This is why the ranking of memory
allocations changes between post-training interventions and matched
training, and why increasing precision after training should be treated
as a change of interface rather than a guaranteed improvement.

The LSTM replication extends the conclusion in two ways. An independently
trained architecture reproduces both the failure under coarse write-back
and its rescue by error feedback, and the state-specific interventions
show that similar numerical suppression has very different consequences
depending on where it occurs: coarse write-back of the cell state is far
more damaging than of the hidden state despite the hidden state's larger
sub-threshold fraction. Because $c$ and $h$ remain coupled, the
single-state interventions are not an additive decomposition, but they
localize the stronger sensitivity to the cell-state interface and show
that the functional role of a stored variable matters alongside its
precision.

Within these bounds, the results change how low-precision recurrent state
should be specified and evaluated. Bit width describes how many values can be
represented, but it does not describe which proposed state changes are
actually written, how long discarded updates can remain absent from the
trajectory, or whether the resulting trajectory is compatible with the
learned recurrent dynamics. A recurrent model and the rule used to store its
state therefore form a coupled temporal system. Evaluating low-precision
recurrent inference requires both the numerical accuracy of the stored state
and the dynamics produced when that state is repeatedly returned to the
network.


\clearpage
\section*{Supplementary material}
\setcounter{figure}{0}
\renewcommand{\thefigure}{S\arabic{figure}}

\setcounter{table}{0}
\renewcommand{\thetable}{S\arabic{table}}

\setcounter{equation}{0}
\renewcommand{\theequation}{S\arabic{equation}}

\setcounter{subsection}{0}
\renewcommand{\thesubsection}{S\arabic{subsection}}

\begin{figure*}[p]
\centering
\includegraphics[width=\textwidth]{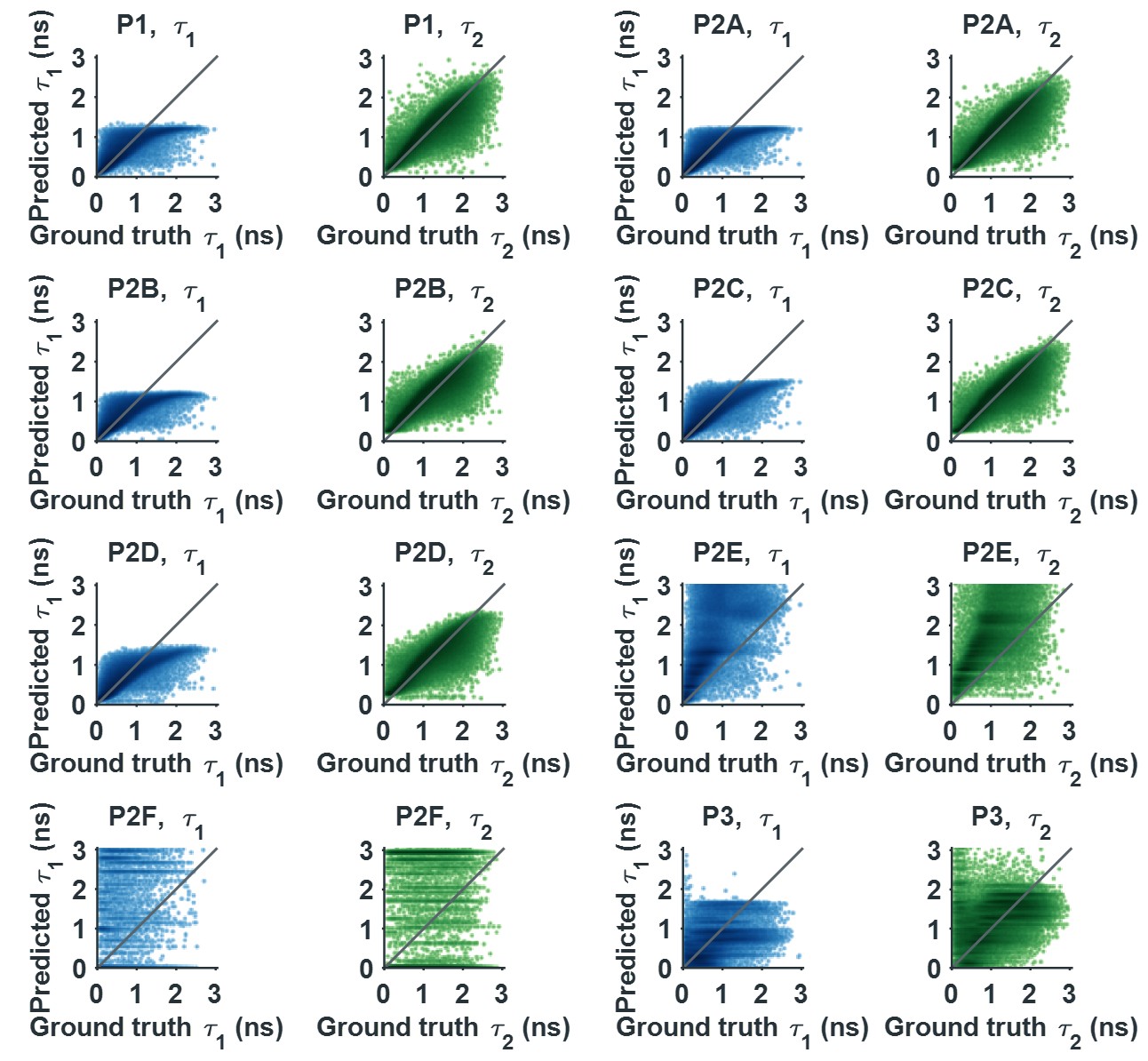}
\caption{\textbf{Phase-resolved lifetime predictions across the QMem
hardening trajectory.}
Predicted-versus-ground-truth lifetime distributions are shown for the
complete 160,000-sample held-out test partition at P1, P2A, P2B, P2C, P2D,
P2E, P2F, and P3. Blue and green density maps represent the local sample
density for $\tau_1$ and $\tau_2$, respectively, and the diagonal line
denotes ideal agreement between prediction and ground truth. The phase sequence shows how task-level behavior evolves as successive
components of the recurrent network are quantized. P2E, in which the
candidate activation is quantized while recurrent-state propagation remains
continuous, shows substantial broadening with predictions beginning to
cluster at discrete output levels. Introducing deterministic 4-bit
recurrent-state write-back at P2F further concentrates predictions at a
small set of discrete lifetime levels and reduces sample-wise correspondence.
Subsequent P3 fine-tuning restores the relationship between predicted and
ground-truth lifetimes. Common axis limits are used within each lifetime parameter to
enable direct comparison across phases; predictions outside the displayed
task range are excluded from the density rendering rather than projected
onto the plot boundaries.}
\label{fig:supp_phase_predictions}
\end{figure*}

\begin{figure*}[p]
\centering
\includegraphics[width=\textwidth]{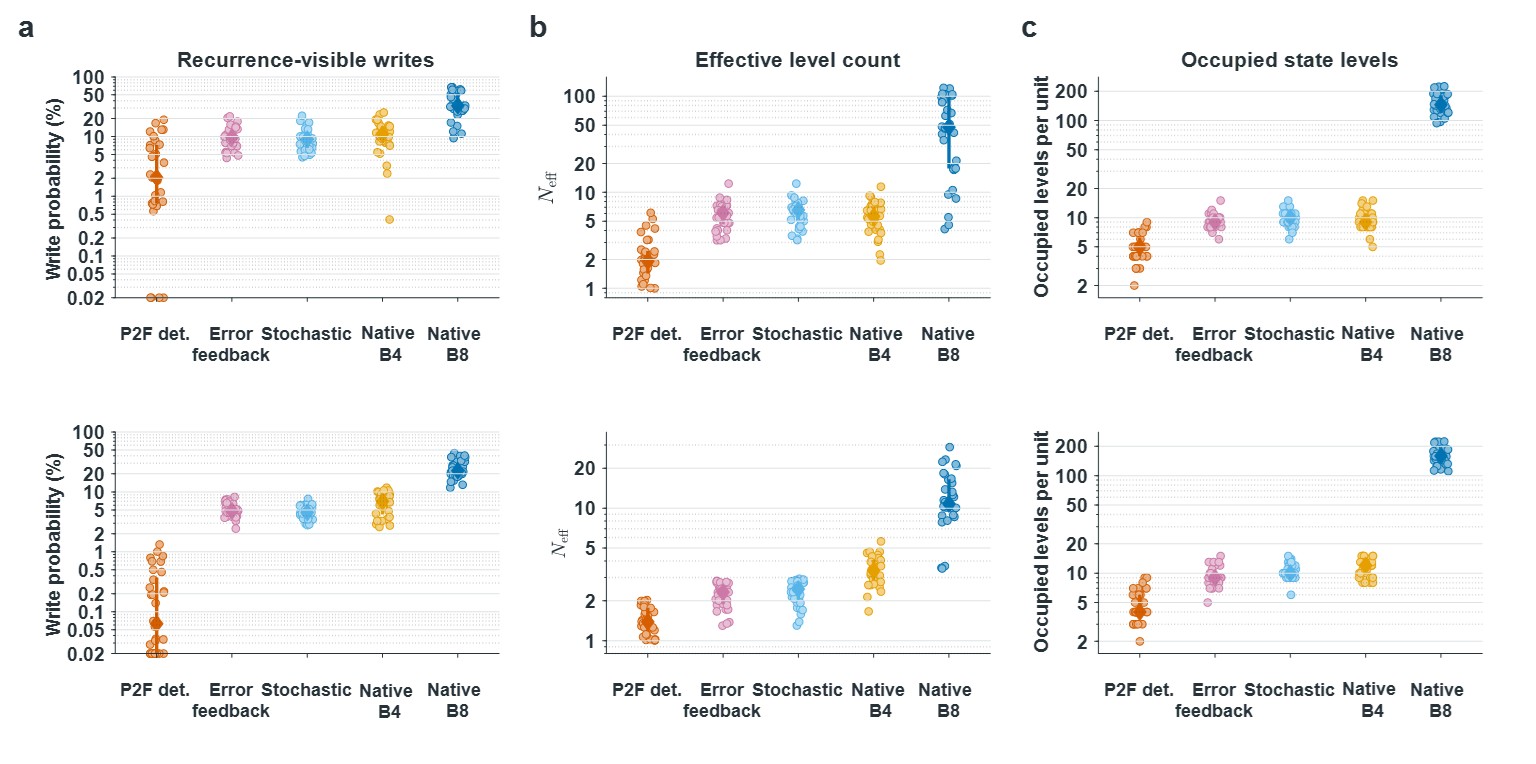}
\caption{\textbf{Per-unit recurrent-state usage across write-back
conditions.}
Distributions across the 32 recurrent hidden units are shown for
deterministic P2F write-back, the fixed P2F error-feedback and
stochastic-rounding interventions, and the native 4-bit and 8-bit reference
models. Encoder and decoder units are shown in the upper and lower rows,
respectively.
\textbf{a,} Recurrence-visible write probability, reported as
$100[1-P(W_0)]$, where $P(W_0)$ is the probability that an individual
stored state element remains unchanged after a recurrent step.
\textbf{b,} Entropy-derived effective level count $N_{\mathrm{eff}}$,
summarizing the distribution of state levels visited by each recurrent
unit.
\textbf{c,} Number of distinct stored state levels occupied by each unit.
Individual circles represent recurrent units, diamonds indicate the median,
and vertical lines span the interquartile range. Logarithmic vertical axes
resolve the broad range of recurrent-state activity and occupancy.
Deterministic P2F write-back produces the strongest suppression of
recurrence-visible decoder writes and the most restricted state-level
usage, whereas error feedback and stochastic rounding increase recurrent
activity and state occupancy at the same fixed checkpoint. Condition colors
are retained consistently across panels and with the main figures.}
\label{fig:supp_per_unit}
\end{figure*}

\begin{table*}[ht!]
\centering
\caption{\textbf{QMem hardening schedule developed for the present study.}
Epochs are configured maxima. ``On'' indicates that the corresponding 4-bit
quantizer is active. P2F introduces recurrent-state quantization through a
15-epoch state blend.}
\label{tab:QMem_schedule}
\scriptsize
\setlength{\tabcolsep}{3.2pt}
\renewcommand{\arraystretch}{1.08}

\resizebox{\textwidth}{!}{%
\begin{tabular}{lcccccccc}
\toprule
Phase
& Epochs
& Candidate
& Reset
& Update
& Biases
& Activation
& State
& LR multiplier \\
\midrule
P1  & 40  & -- & -- & -- & -- & -- & -- & 1.0 \\
P2A & 30  & On & -- & -- & -- & -- & -- & 0.5 \\
P2B & 30  & On & On & -- & -- & -- & -- & 0.5 \\
P2C & 10  & On & On & On & -- & -- & -- & 0.5 \\
P2D & 10  & On & On & On & On & -- & -- & 0.3 \\
P2E & 30  & On & On & On & On & On & -- & 0.3 \\
P2F & 30  & On & On & On & On & On & On & 0.3 \\
P3  & 170 & On & On & On & On & On & On & 0.1 \\
\bottomrule
\end{tabular}%
}
\end{table*}

\section*{Supplementary Results}
\begin{figure*}[t]
\centering
\includegraphics[width=\textwidth]{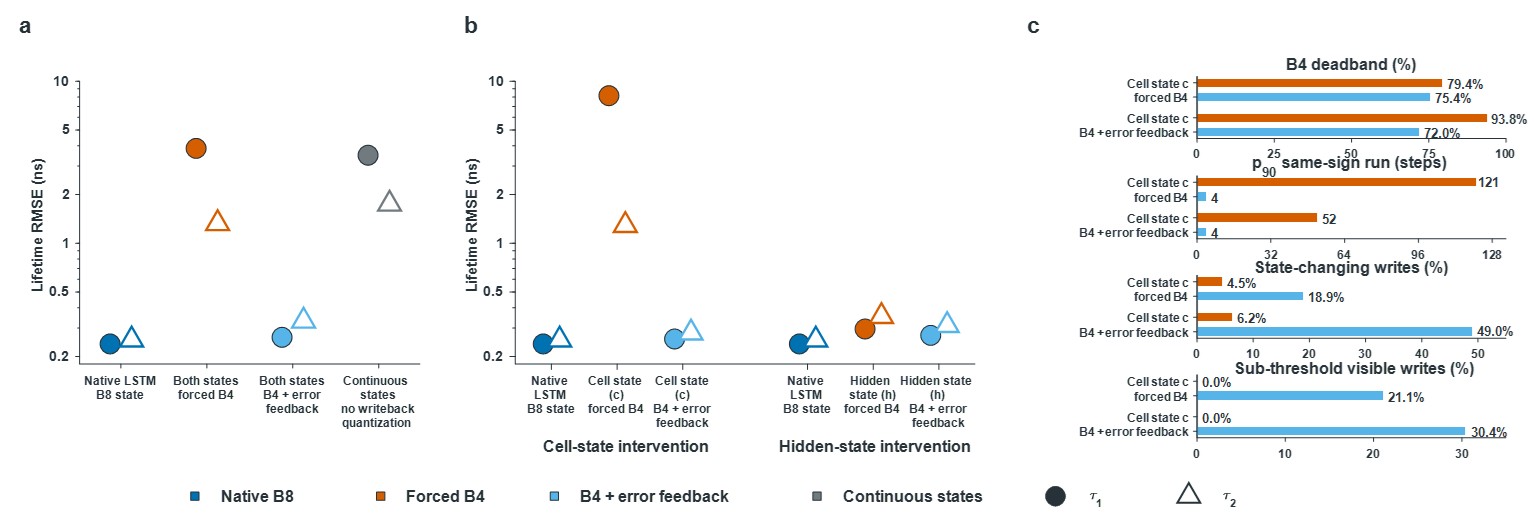}
\caption{\textbf{An independently trained LSTM reproduces the write-back
failure and reveals state-specific sensitivity.}
\textbf{a,} Lifetime RMSE under native 8-bit write-back, deterministic 4-bit
write-back on both recurrent states, 4-bit error feedback on both states,
and continuous propagation.
\textbf{b,} Interventions targeting only the cell state $c$ or only the
hidden state $h$ (deterministic 4-bit or 4-bit error feedback) while the
other state stays at 8-bit.
\textbf{c,} Decoder diagnostics for the targeted state: fraction of proposed
changes inside the 4-bit write boundary, 90th-percentile same-direction run
length, state-changing writes, and sub-threshold writes that become visible.
Circles: $\tau_1$; triangles: $\tau_2$; RMSE in nanoseconds, logarithmic
axes.}
\label{fig:lstm_replication}
\end{figure*}

\subsection{Location and lifetime-conditioned controls}
\label{sec:supp_controls}
The frozen P2E location control separates sensitivity of the encoder and
decoder recurrent regions. Applying 4-bit write-back only to the encoder gives
$\tau_1$ and $\tau_2$ RMSE values of 2.55 and 1.12~ns, while decoder-only 4-bit
write-back gives 7.79 and 8.36~ns. Applying 4-bit write-back to both regions
gives 39.04 and 24.58~ns (Supplementary Table~\ref{tab:location_control}). The decoder is therefore more sensitive than the
encoder in this frozen checkpoint, while the combined intervention shows
that both recurrent regions contribute to the complete sequence-to-sequence
trajectory. The P2F operator effect is also present throughout the evaluated
lifetime range. Deterministic 4-bit write-back increases RMSE relative to
same-checkpoint identity propagation in all ten equal-count ground-truth
bins for both $\tau_1$ and $\tau_2$, with the paired 95\% confidence
interval above zero in every bin. The excess RMSE ranges from 18.86 to
39.42~ns across the $\tau_1$ bins and from 58.18 to 159.02~ns across the
$\tau_2$ bins. The magnitude is not monotonic with ground-truth lifetime, so
the present measurements do not support a simple interpretation in which
longer lifetime values are uniquely sensitive to coarse state write-back.

\begin{table}[ht!]
\centering
\caption{P2E location control. The P2E checkpoint is held fixed and 4-bit state
write-back is applied only to the indicated recurrent region.}
\label{tab:location_control}
\scriptsize
\setlength{\tabcolsep}{5.0pt}
\renewcommand{\arraystretch}{1.10}

\begin{tabular}{lccc}
\toprule
Condition & Sequence MAE & $\tau_1$ RMSE & $\tau_2$ RMSE \\
\midrule
P2E native           & 0.139248 & 1.3975  & 3.0309 \\
Encoder-only 4-bit      & 0.071919 & 2.5538  & 1.1181 \\
Decoder-only 4-bit      & 0.544145 & 7.7854  & 8.3607 \\
Encoder + decoder 4-bit & 0.105206 & 39.0445 & 24.5779 \\
\bottomrule
\end{tabular}
\end{table}

\subsection{Sequence reconstruction and lifetime error}
\label{sec:supp_metric_interpretation}

Sequence MAE and lifetime RMSE do not necessarily preserve the same ranking
under write-limited recurrence. For example, deterministic P2F has sequence
MAE 0.09 despite substantially larger lifetime errors than same-checkpoint
identity propagation. Lifetime is derived from the temporal structure of the
predicted decay, so a comparatively small pointwise reconstruction error can
coexist with a decay shape that gives inaccurate integrated lifetime.
Lifetime RMSE is therefore treated as the primary task-level metric, with
sequence MAE retained as a complementary reconstruction measure.

\subsection{LSTM cross-architecture and state-type controls}
\label{sec:supp_lstm_results}

The LSTM analysis provides a frozen cross-architecture test using a
checkpoint whose reconstructed native outputs are exactly equivalent to the
original implementation on the complete held-out test set. Whole-state
interventions test whether the GRU write-back effect is reproduced when both
LSTM recurrent states are changed together. State-type interventions then
change only $c$ or only $h$ while retaining the other state at native 8-bit
write-back. Complete task-level metrics are reported in Supplementary
Table~\ref{tab:lstm_frozen_performance}. State diagnostics for the recurrent
variable targeted by each intervention are reported in Supplementary
Table~\ref{tab:lstm_state_diagnostics}.

\begin{table*}[ht!]
\centering
\caption{\textbf{Post-training LSTM recurrent-state write-back interventions.}
The independently trained LSTM checkpoint is fixed in every row. Native
inference uses deterministic 8-bit write-back for both $c$ and $h$. 4-bit
error-feedback rows retain a 4-bit recurrence-visible state while preserving
discarded error in the auxiliary error-feedback state. Identity denotes
continuous state propagation without discrete recurrent-state write-back.
All metrics are evaluated on the complete 160,000-sample held-out test partition. Lifetime RMSE is reported in nanoseconds, and
$r_{\tau_1}$ and $r_{\tau_2}$ denote the Pearson correlation between
predicted and ground-truth lifetimes across test samples.}
\label{tab:lstm_frozen_performance}
\scriptsize
\setlength{\tabcolsep}{3.2pt}
\renewcommand{\arraystretch}{1.08}

\resizebox{\textwidth}{!}{%
\begin{tabular}{lllccccc}
\toprule
Condition
& $c$ write-back
& $h$ write-back
& Sequence MAE
& $r_{\tau_1}$
& $\tau_1$ RMSE
& $r_{\tau_2}$
& $\tau_2$ RMSE \\
\midrule

Native
& Deterministic 8-bit
& Deterministic 8-bit
& 0.016952
& 0.856660
& 0.239467
& 0.892168
& 0.253627 \\

Both states 4-bit
& Deterministic 4-bit
& Deterministic 4-bit
& 0.106844
& -0.014412
& 3.857162
& 0.049316
& 1.327105 \\

Both states 4-bit + error feedback
& Error feedback 4-bit
& Error feedback 4-bit
& 0.034085
& 0.847885
& 0.263191
& 0.864387
& 0.332004 \\

Identity propagation
& Identity
& Identity
& 0.209483
& 0.516141
& 3.500654
& 0.839935
& 1.748782 \\

$c$-only 4-bit
& Deterministic 4-bit
& Deterministic 8-bit
& 0.111953
& -0.010161
& 8.158223
& -0.041065
& 1.291688 \\

$c$-only 4-bit + error feedback
& Error feedback 4-bit
& Deterministic 8-bit
& 0.028628
& 0.848464
& 0.256951
& 0.870164
& 0.280138 \\

$h$-only 4-bit
& Deterministic 8-bit
& Deterministic 4-bit
& 0.040708
& 0.781670
& 0.296195
& 0.785545
& 0.354278 \\

$h$-only 4-bit + error feedback
& Deterministic 8-bit
& Error feedback 4-bit
& 0.027088
& 0.852056
& 0.270545
& 0.876796
& 0.311550 \\

\bottomrule
\end{tabular}%
}
\end{table*}

\begin{table*}[p]
\centering
\caption{\textbf{LSTM recurrent-state diagnostics for native and targeted
write-back conditions.}
Statistics are reported separately for the recurrent state targeted by each
intervention. Deadband fraction is the fraction of proposed state changes
within the deterministic half-step boundary. State-change fraction is the
fraction of recurrence-visible state elements that change between
successive recurrent steps. $P_{\mathrm{sub}\rightarrow\mathrm{write}}$ is
the conditional fraction of sub-threshold proposed state changes that nevertheless
produce a recurrence-visible transition. $P_{\mathrm{same}}$ is the
same-sign fraction over adjacent eligible proposed state change pairs. Run lengths are
completed same-sign runs in recurrent write steps.}
\label{tab:lstm_state_diagnostics}
\scriptsize
\setlength{\tabcolsep}{2.8pt}
\renewcommand{\arraystretch}{1.08}

\resizebox{\textwidth}{!}{%
\begin{tabular}{lllcccccc}
\toprule
Condition
& Region
& State
& Deadband fraction
& State-change fraction
& $P_{\mathrm{sub}\rightarrow\mathrm{write}}$
& $P_{\mathrm{same}}$
& Median run
& $p_{90}$ run \\
\midrule

Native
& Encoder
& $c$
& 0.061555
& 0.361288
& 0.000000
& 0.742438
& 1
& 3 \\

Native
& Decoder
& $c$
& 0.352734
& 0.214999
& 0.000000
& 0.940647
& 2
& 18 \\

Native
& Encoder
& $h$
& 0.368190
& 0.631703
& 0.000000
& 0.700942
& 1
& 3 \\

Native
& Decoder
& $h$
& 0.782909
& 0.217015
& 0.000000
& 0.948779
& 1
& 15 \\

\midrule

$c$-only 4-bit
& Encoder
& $c$
& 0.570220
& 0.049930
& 0.000000
& 0.968113
& 4
& 24 \\

$c$-only 4-bit
& Decoder
& $c$
& 0.794329
& 0.044821
& 0.000000
& 0.991887
& 4
& 121 \\

$c$-only 4-bit + error feedback
& Encoder
& $c$
& 0.424069
& 0.113641
& 0.144799
& 0.916727
& 2
& 7 \\

$c$-only 4-bit + error feedback
& Decoder
& $c$
& 0.754240
& 0.189500
& 0.211226
& 0.860292
& 1
& 4 \\

\midrule

$h$-only 4-bit
& Encoder
& $h$
& 0.867134
& 0.109326
& 0.000000
& 0.900874
& 2
& 11 \\

$h$-only 4-bit
& Decoder
& $h$
& 0.937767
& 0.062055
& 0.000000
& 0.973244
& 3
& 52 \\

$h$-only 4-bit + error feedback
& Encoder
& $h$
& 0.702117
& 0.463023
& 0.289989
& 0.760958
& 1
& 4 \\

$h$-only 4-bit + error feedback
& Decoder
& $h$
& 0.719932
& 0.490272
& 0.304164
& 0.801690
& 2
& 4 \\

\bottomrule
\end{tabular}%
}
\end{table*}

\subsection{Post-training recurrent-state precision sweeps}
\label{sec:supp_precision_sweeps}
\begin{figure*}[t]
\centering
\includegraphics[width=\textwidth]{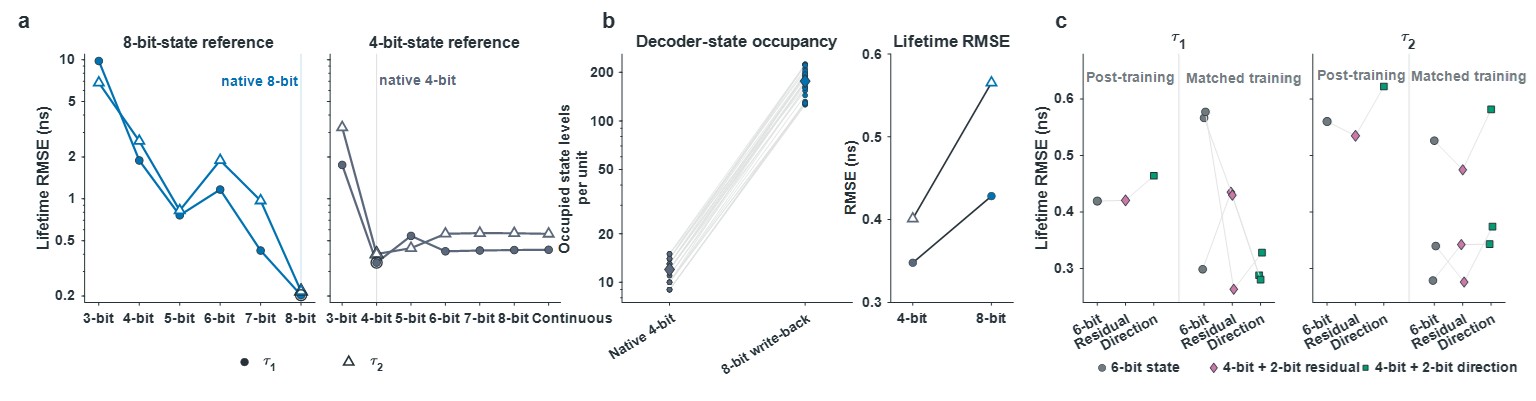}
\caption{\textbf{State precision is not a monotonic control variable, and
recurrent-memory compatibility is learned.}
\textbf{a,} Post-training state-precision sweeps for independently trained
8-bit-state and 4-bit-state reference GRUs; only the state write-back rule
changes while trained parameters stay fixed. The 4-bit-state sweep also
includes continuous propagation; native precision is marked. Circles:
$\tau_1$; triangles: $\tau_2$.
\textbf{b,} Decoder-state occupancy and lifetime error for the 4-bit-state
reference GRU under native 4-bit versus 8-bit write-back; lines pair the
same decoder unit. Median occupied levels rise from 12 to 176.5 while RMSE
worsens from 0.35/0.40~ns to 0.43/0.57~ns.
\textbf{c,} Six stored bits per unit allocated as a 6-bit state, a 4-bit
state plus 2-bit residual memory, or a 4-bit state plus 2-bit direction
memory, compared under post-training intervention and matched training;
lines connect matched training runs.}
\label{fig:state_precision_alignment}
\end{figure*}

Post-training state-precision sweeps provide the complementary test of whether
numerical resolution alone determines recurrent performance
(Fig.~\ref{fig:state_precision_alignment}a). The 8-bit-state reference GRU reaches
0.20 and 0.22~ns at its native 8-bit state representation and becomes
substantially less accurate on coarser grids, but performance is not
monotonically ordered by state bit width across the sweep. The independently
trained 4-bit model shows the opposite constraint: it reaches 0.35 and 0.40~ns
at native 4-bit, while 8-bit state propagation gives 0.43 and 0.57~ns and
continuous-state propagation gives 0.43 and 0.56~ns. Thus, additional state
precision does not necessarily move a fixed recurrent solution toward
better task performance.

The increased numerical resolution is nevertheless used by the recurrent
state. When the frozen 4-bit-state reference GRU is changed from native 4-bit to 8-bit
state write-back, the median number of occupied decoder levels rises from 12
to 176.5, with the increase present across all 32 decoder units
(Fig.~\ref{fig:state_precision_alignment}b). Despite this large expansion of
the accessible state trajectory, lifetime RMSE worsens rather than improves.
The failure of finer state resolution to improve the task is therefore not
explained by the network simply ignoring the additional representational
levels. Complete precision sweeps are reported in Supplementary
Table~\ref{tab:supp_precision_sweeps}.

\begin{table*}[ht!]
\centering
\caption{\textbf{Post-training recurrent-state precision sweeps.} Trained weights
remain fixed and only recurrent-state write-back precision is changed. RMSE
is reported in nanoseconds.}
\label{tab:supp_precision_sweeps}
\scriptsize
\setlength{\tabcolsep}{5.0pt}
\renewcommand{\arraystretch}{1.10}
\resizebox{0.92\textwidth}{!}{
\begin{tabular}{llccc}
\toprule
Trained model & State write-back & Sequence MAE & $\tau_1$ RMSE & $\tau_2$ RMSE \\
\midrule
Native 8-bit & 3-bit & 0.109479 & 9.7973 & 6.8297 \\
Native 8-bit & 4-bit & 0.102485 & 1.8873 & 2.6030 \\
Native 8-bit & 5-bit & 0.070121 & 0.7610 & 0.8244 \\
Native 8-bit & 6-bit & 0.124992 & 1.1659 & 1.8921 \\
Native 8-bit & 7-bit & 0.053859 & 0.4246 & 0.9715 \\
Native 8-bit & 8-bit & 0.014685 & 0.2028 & 0.2164 \\
\midrule
Native 4-bit & 3-bit & 0.160885 & 1.7527 & 3.2619 \\
Native 4-bit & 4-bit & 0.028412 & 0.3479 & 0.4011 \\
Native 4-bit & 5-bit & 0.034009 & 0.5418 & 0.4418 \\
Native 4-bit & 6-bit & 0.037540 & 0.4195 & 0.5604 \\
Native 4-bit & 7-bit & 0.038162 & 0.4251 & 0.5671 \\
Native 4-bit & 8-bit & 0.038061 & 0.4284 & 0.5653 \\
Native 4-bit & Float & 0.038022 & 0.4300 & 0.5602 \\
\bottomrule
\end{tabular}
}
\end{table*}

\subsection{Recurrent-memory allocation under matched storage}
\label{sec:supp_matched_storage}
The fixed 4-bit-state reference checkpoint provides an additional comparison of how
stored recurrent-memory bits are allocated while trained parameters remain
fixed. Deterministic 6-bit, 7-bit, and 8-bit state write-back assigns all additional
bits to the recurrence-visible state. Alternative conditions retain the 4-bit
recurrence-visible state and allocate the corresponding additional bits to a
quantized residual memory or direction memory. At six stored bits per unit, the
comparison is mixed. At seven and eight bits, several auxiliary-memory
configurations produce lower lifetime RMSE than assigning all additional
bits to recurrence-visible state precision. At eight stored bits per unit, a 4-bit recurrence-visible state with 4-bit
direction memory reaches 0.36 and 0.46~ns, compared with 0.43 and 0.57~ns
for 8-bit state write-back on the same fixed 4-bit-state reference GRU (Supplementary Table~\ref{tab:equal_storage}). Native
deterministic 4-bit itself remains more accurate than either condition. This
comparison is therefore interpreted as an operator-compatibility result, not
as evidence that auxiliary bits are intrinsically more valuable than state
bits. Residual and direction-memory storage influence when future 4-bit transitions
occur while retaining the recurrence-visible 4-bit interface. In contrast,
additional state bits directly change the values presented to the recurrent
kernel. The comparison matches stored recurrent-memory bits per hidden unit
but does not imply equal arithmetic, logic, routing, energy, or hardware
cost.

\begin{table}[ht!]
\centering
\caption{\textbf{Matched recurrent-memory storage for the fixed 4-bit-state
reference checkpoint.} At each stored bit count, a 4-bit recurrence-visible
state with $k$ auxiliary bits is compared with deterministic $(4+k)$-bit
state write-back that
assigns all bits to the visible state (baseline). $\Delta$ is alternative
minus baseline lifetime RMSE in nanoseconds; negative values favor the
auxiliary-memory condition. Paired sequence-level 95\% bootstrap intervals (2,000 replicates) have
half-widths below $0.0012$~ns. Native deterministic 4-bit on the same checkpoint gives
0.348/0.401~ns. The comparison matches stored recurrent bits per hidden
unit and does not imply equal hardware cost.}
\label{tab:equal_storage}
\small
\setlength{\tabcolsep}{5pt}
\renewcommand{\arraystretch}{1.10}
\begin{tabular}{clcccc}
\toprule
Bits & Alternative & $\tau_1$ RMSE & $\Delta\mathrm{RMSE}_{\tau_1}$ & $\tau_2$ RMSE & $\Delta\mathrm{RMSE}_{\tau_2}$ \\
\midrule
6 & Baseline 6-bit                          & 0.420 & --       & 0.560 & --       \\
  & 4-bit + R2                              & 0.421 & $+0.001$ & 0.535 & $-0.025$ \\
  & 4-bit state + 2-bit direction memory     & 0.464 & $+0.045$ & 0.622 & $+0.062$ \\
\midrule
7 & Baseline 7-bit                          & 0.425 & --       & 0.567 & --       \\
  & 4-bit + R3                              & 0.423 & $-0.002$ & 0.544 & $-0.023$ \\
  & 4-bit state + 3-bit direction memory     & 0.398 & $-0.027$ & 0.510 & $-0.057$ \\
\midrule
8 & Baseline 8-bit                          & 0.428 & --       & 0.565 & --       \\
  & 4-bit + R4                              & 0.421 & $-0.008$ & 0.538 & $-0.027$ \\
  & 4-bit state + 4-bit direction memory     & 0.359 & $-0.069$ & 0.464 & $-0.101$ \\
\bottomrule
\end{tabular}
\end{table}

\subsection{Matched training around the recurrent-state interface}
\label{sec:supp_matched_training_results}
Four recurrent-memory allocations were trained from matched initializations
under the campaign described in Supplementary
Methods~\ref{sec:supp_trained_campaign}: 4-bit state, 6-bit state, 4-bit
state with 2-bit residual memory, and 4-bit state with 2-bit direction
memory. Three matched training runs were evaluated for each condition. On
$\tau_1$, direction memory reaches mean RMSE $0.30\pm0.03$~ns, compared
with $0.41\pm0.03$ for 4-bit state, $0.48\pm0.16$ for 6-bit state, and
$0.38\pm0.10$ for residual memory. Direction memory improves on the 4-bit
and 6-bit-state conditions in all three matched runs. On $\tau_2$, the
6-bit-state and residual-memory conditions improve on 4-bit state in all
three matched runs, whereas direction memory is higher than both in all
three. Complete run-level values and paired differences are reported in
Supplementary Tables~\ref{tab:trained_campaign} and
\ref{tab:trained_campaign_paired}. The campaign is therefore used to
establish learned interface compatibility rather than a universal ranking
of recurrent-memory operators.

\clearpage
\section*{Supplementary Methods for Reproducibility}
\label{sec:Supplementarymethods}

These supplementary methods document the analysis path used to connect phase-wise lifetime behavior with recurrent-state dynamics. The same fixed dataset, held-out evaluation partition, lifetime extraction procedure, and
write-back definitions are retained across the frozen analyses. The primary mechanistic study uses the Seq2SeqLite GRU checkpoints, while an independently trained LSTM checkpoint provides the cross-architecture frozen replication and state-type dissection. 
Details are separated here from the main narrative so that the diagnostic interpretation can be traced from training configuration through state-grid analysis without altering the reported trained checkpoints.

\subsection{Training configuration and model architecture}
\label{sec:supp_training}

\subsubsection*{Model and dataset provenance}

The experiments use the Seq2Seq framework for time-resolved fluorescence
reconstruction and lifetime estimation described in
\cite{pandey2024deep}. Seq2Seq uses a two-layer 128-unit GRU
encoder--decoder to transform a 135-step time-resolved fluorescence signal
into temporal output parameters from which fluorescence parameters are
estimated. Seq2SeqLite is the compressed student architecture introduced in
\cite{erbas2024compressing,erbas2024unlocking}, consisting of a single-layer
32-unit GRU encoder--decoder with 6,627 trainable parameters.

The present analysis uses the published simulation pipeline, fixed
train/validation/test partition, 128-unit teacher architecture, and 32-unit
Seq2SeqLite student configuration so that recurrent-state interventions are
evaluated within a common fluorescence-lifetime inference setting. QMem is
introduced in the present study as a staged quantization-aware training
procedure that progressively applies low-precision constraints to
Seq2SeqLite while optimization continues. No additional synthetic dataset is
generated for the QMem experiments.

Signals were generated with the PyFLI simulation framework using mono- and
bi-exponential fluorescence kinetics, photon statistics, detector-dependent
noise, and experimentally measured pixel-wise instrument response functions
(IRFs), as described in
\cite{pandeypyfli,erbas2024compressing}. For sample $i$, the
bi-exponential fluorescence impulse response is
\begin{equation}
f_i(t)
=
I_i
\left[
a_i
\exp\left(-\frac{t}{\tau_{1,i}}\right)
+
(1-a_i)
\exp\left(-\frac{t}{\tau_{2,i}}\right)
\right]
u(t),
\label{eq:supp_biexponential_decay}
\end{equation}
where
$\boldsymbol{\theta}_i=(\tau_{1,i},\tau_{2,i})$ is the lifetime-parameter
vector introduced in the main text,
$a_i\in[0,1]$ is the fractional contribution of the short-lifetime
component, $I_i>0$ is the signal scale, and $u(t)$ is the unit-step function
enforcing a causal decay.

For excitation period $T$, the ideal periodic excitation train is
\begin{equation}
x_T(t)
=
\sum_{m\in\mathbb{Z}}
\delta_{\mathrm{D}}(t-mT),
\qquad
T=12.5~\mathrm{ns},
\label{eq:supp_excitation_train}
\end{equation}
where $\delta_{\mathrm{D}}(t)$ denotes the Dirac impulse. Let $E_i(t)$ denote the temporal
response of the acquisition chain at sample or pixel $i$ to an
instantaneous excitation. The corresponding periodic instrument response
function is
\begin{equation}
\mathrm{IRF}_{i,T}(t)
=
\left(E_i*x_T\right)(t).
\label{eq:supp_irf}
\end{equation}
The recorded time-resolved fluorescence signal can then be written as
\begin{align}
y_i(t)
&=
\left[
E_i*
\left(
x_T*f_i
\right)
\right](t)
+
\eta_i(t)
\\
&=
\left[
\left(
E_i*x_T
\right)
*f_i
\right](t)
+
\eta_i(t)
\\
&=
\left[
\mathrm{IRF}_{i,T}
*f_i
\right](t)
+
\eta_i(t),
\label{eq:supp_observation_model}
\end{align}
where $*$ denotes convolution and $\eta_i(t)$ represents the simulated
photon-statistics and detector-dependent noise contributions. Thus, the
network input is a high-noise time-resolved fluorescence signal whose
temporal structure reflects both the fluorescence kinetics and the
instrument response. Each sample additionally contains the three-parameter
temporal target used by the decoder loss and scalar labels for
$\tau_{1,i}$, $\tau_{2,i}$, and $a_i$. The dataset contains 1,600,000 simulated time-resolved fluorescence signals, each represented by 135 temporal bins. A fixed 80/10/10 partition provides 1,280,000 training samples, 160,000 validation samples, and 160,000 held-out test samples. The held-out partition is not used for training, checkpoint selection, learning-rate adaptation, or generation of distillation targets. The teacher, native quantized controls, and every QMem phase use the same partition, ensuring that all reported comparisons are evaluated under common data conditions.

\subsubsection*{Teacher and student architectures}

The frozen teacher is the two-layer GRU encoder--decoder from \cite{erbas2024compressing, pandey2024deep}, with 128 hidden units in each recurrent layer. Encoder final states initialize the decoder, which receives a zero-valued input sequence. A linear dense layer then maps the decoder hidden trajectory to the three-parameter output sequence.

The student is Seq2SeqLite, a single-layer GRU encoder--decoder with 32 hidden units and a linear dense readout. The encoder processes the normalized 135-step input sequence, and its final hidden state initializes a
decoder receiving a 135-step zero input. The resulting 32-unit architecture contains 6,627 trainable parameters and corresponds to the compact student family selected in the design-space evaluation of \cite{erbas2024compressing, erbas2024unlocking}.

For QMem, the Phase-2 training graph exposes the decoder hidden trajectory and pre-sigmoid update-gate logits alongside the predicted sequence. These diagnostic outputs add no trainable parameters. After Phase 2, the separated
gate matrices are repacked into the standard GRU inference graph used for final P3 inference, preserving the intended deployment graph.

\subsubsection*{Knowledge distillation and optimization}
Predictions from the frozen teacher checkpoint are generated once and reused as fixed distillation targets. Student optimization then minimizes a combination of supervised sequence error and teacher--student sequence
error,

\begin{equation}
\mathcal{L}_{\mathrm{base}}
=
(1-\alpha)\mathcal{L}_{\mathrm{task}}
+
\alpha\mathcal{L}_{\mathrm{KD}},
\label{eq:supp_base_loss}
\end{equation}
where
\begin{align}
\mathcal{L}_{\mathrm{task}}
&=
\frac{1}{NTC}
\sum_{i=1}^{N}
\sum_{t=1}^{T}
\sum_{c=1}^{C}
\left(
\hat{y}^{S}_{i,t,c}
-
y_{i,t,c}
\right)^2, \\
\mathcal{L}_{\mathrm{KD}}
&=
\frac{1}{NTC}
\sum_{i=1}^{N}
\sum_{t=1}^{T}
\sum_{c=1}^{C}
\left(
\hat{y}^{S}_{i,t,c}
-
\hat{y}^{T}_{i,t,c}
\right)^2 .
\end{align}
Here $N$ is the number of samples, $T$ is the number of temporal bins,
$C$ is the number of output parameters, and superscripts $S$ and $T$ denote
student and teacher predictions, respectively.

The distillation weight is fixed at $\alpha=0.6$, following the
Seq2SeqLite compression configuration
\cite{erbas2024compressing}. The regression knowledge-distillation objective is direct mean-squared error
between the student and teacher sequence outputs.

Optimization uses Adam with batch size 1024 and base learning rate
$10^{-4}$. Phase 1 uses a learning-rate multiplier of 1.0; P2A--P2C use 0.5;
P2D--P2F use 0.3; and P3 uses 0.1, giving initial phase rates of $10^{-4}$, $5\times10^{-5}$, $3\times10^{-5}$, and $10^{-5}$, respectively. 
Phase 1 includes a five-epoch linear warm-up, whereas later phases begin directly at their assigned rates. After eight epochs without the required validation improvement, the learning rate is reduced by 0.5 to a minimum of $10^{-6}$;
P3 additionally uses a phase-specific floor of $5\times10^{-6}$. The
improvement threshold is $10^{-5}$ and early-stopping patience is 30 epochs. Training uses float32 throughout, with mixed precision disabled.

\subsubsection*{Quantized GRU formulation}

The QMem configuration uses 4-bit input kernels, recurrent kernels, biases,
and candidate activation and, after recurrent-state quantization is
introduced at P2F, 4-bit recurrent-state storage. Kernels, recurrent kernels, biases, and recurrent states use uniform
quantization, and the candidate activation uses a symmetric quantized
hyperbolic tangent \cite{qkeras}. For the
recurrent-state grid used in the 4-bit analyses,
\begin{equation}
\Delta_4=0.125,
\end{equation}
with half-step $\Delta_4/2=0.0625$.

The complete GRU recurrence and state-write-back notation are defined in
Supplementary Methods~\ref{sec:supp_write-back}.

\subsubsection*{QMem progressive quantization procedure}
\begin{figure*}[ht!]
\centering
\includegraphics[width=\textwidth]{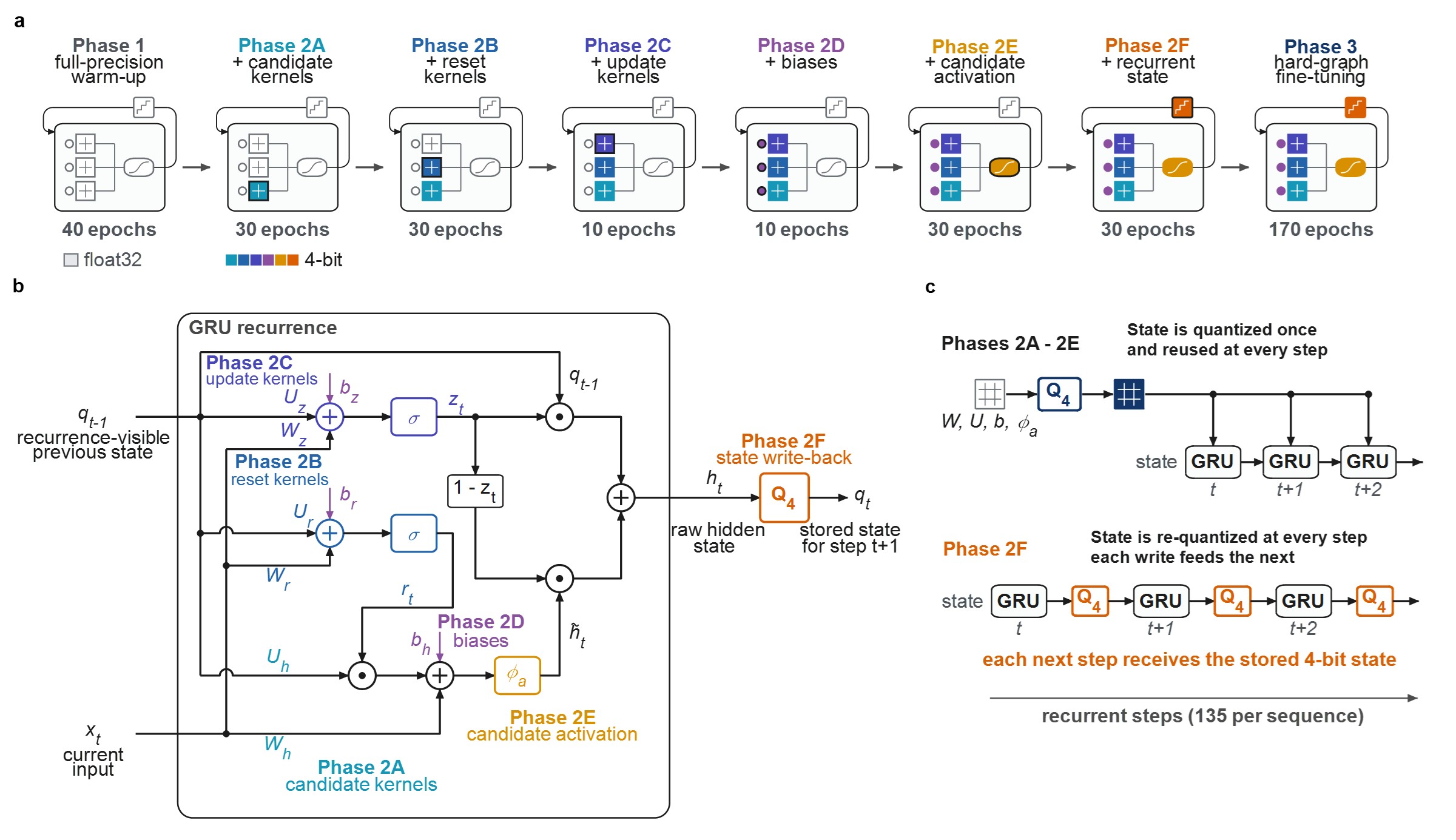}
\caption{\textbf{QMem progressively hardens the GRU to 4-bit operation with
recurrent-state quantization introduced last.}
\textbf{a,} QMem training sequence. Starting from full precision (P1),
4-bit quantization is introduced sequentially for candidate kernels (P2A),
reset-gate kernels (P2B), update-gate kernels (P2C), biases (P2D),
candidate activation (P2E), and recurrent state (P2F). Each quantizer remains
active in subsequent phases. P2F progressively moves the recurrent state onto the 4-bit grid over its
first 15 epochs, after which deterministic 4-bit state write-back is
retained. P3 continues training after the fully quantized GRU is repacked into the standard inference
graph. Numbers below each phase give the configured maximum epochs
(Supplementary Table~\ref{tab:QMem_schedule}).
\textbf{b,} Implemented GRU recurrence with the QMem quantization targets
located on the corresponding operations. The recurrent state is stored through
$Q_4$ and returned as $q_t$ to the next recurrent step.
\textbf{c,} Unlike kernels, biases, and activation parameters, whose quantized
values are reused across the sequence, the recurrent state is written again
after every step. Each stored state therefore conditions the following
recurrent computation, motivating the frozen write-back analyses in the main
text.}
\label{fig:supp_qmem}
\end{figure*}
QMem was developed for the present study as a staged
quantization-hardening procedure. It begins with a floating-point warm-up,
proceeds through six Phase-2 hardening stages, and ends with a final
hard-graph polishing stage (Supplementary Table~\ref{tab:QMem_schedule}). A
standard GRU stores gate parameters in packed matrices. For diagnostic
training, QMem separates candidate, reset, and update input and recurrent
kernels so that each pathway can be hardened independently. This split
representation is used only for analysis; after P2F, the parameters are
repacked into the standard inference graph. The corresponding phase-resolved
predicted-versus-ground-truth lifetime distributions on the common held-out
test partition are shown in Supplementary
Fig.~\ref{fig:supp_phase_predictions}.

The best validation checkpoint from each phase initializes the next phase;
optimizer state is not transferred. The corresponding training and
validation trajectory across the complete hardening sequence is shown in
Supplementary Fig.~\ref{fig:supp_training_history}. P2E uses hard 4-bit candidate activation while recurrent
state remains continuous. P2F introduces recurrent-state quantization progressively over its first
15 epochs. Full 4-bit
state write-back is reached at the fifteenth P2F epoch and is retained
thereafter. Following P2F, the separated gate matrices are repacked into the
standard QKeras GRU graph used for P3.

\begin{figure*}[t]
\centering
\includegraphics[width=\textwidth]{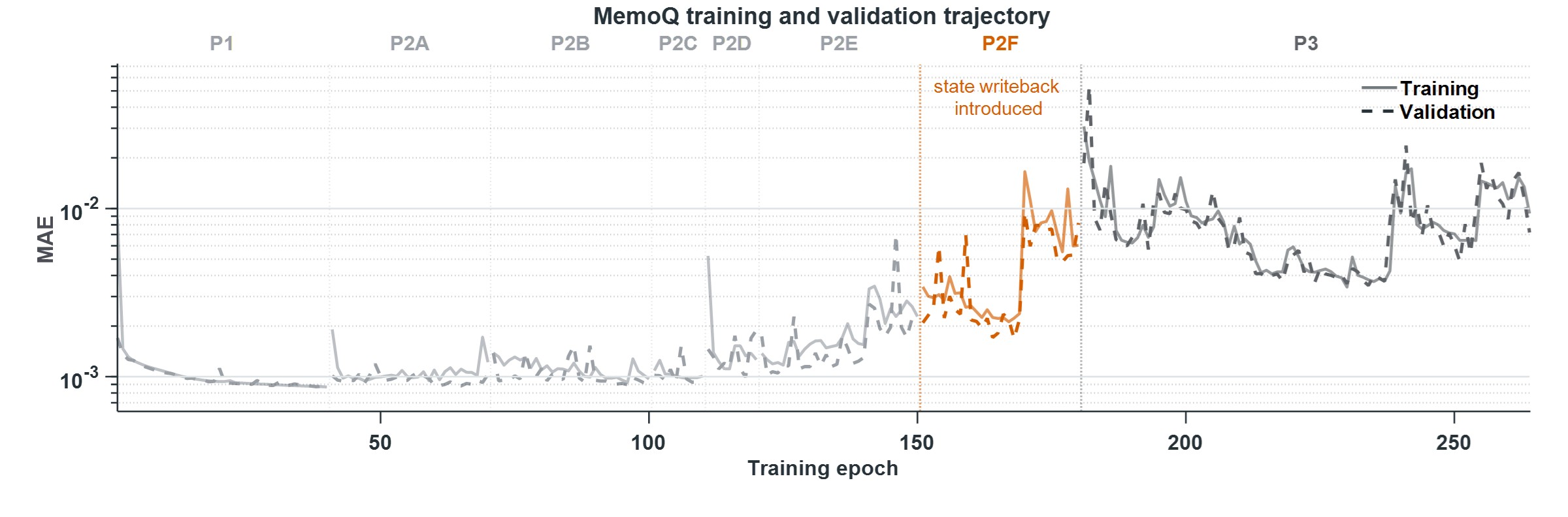}
\caption{\textbf{Training and validation dynamics across the QMem
hardening trajectory.}
Training and validation mean absolute error (MAE) recorded during the
primary QMem trajectory are shown across P1, P2A, P2B, P2C, P2D, P2E, P2F,
and P3. Solid and dashed curves denote training and validation values,
respectively, and vertical dotted lines indicate transitions between
successive hardening phases. P1 through P2E are shown in neutral gray, P2F
is highlighted in orange at the introduction of recurrent-state
quantization, and P3 is shown in graphite for the final repacked hard graph.
Each phase is initialized from the best validation checkpoint of the
preceding phase, with optimizer state reset between phases. Curves are
plotted directly from the saved training history without smoothing.}
\label{fig:supp_training_history}
\end{figure*}

\subsection{Post-training write-back interventions and diagnostics}
\label{sec:supp_write-back}

Post-training intervention analysis uses saved checkpoints without retraining
or fine-tuning. The encoder, decoder, and dense readout are reconstructed
from the trained parameters and first evaluated using the state-storage rule
associated with the original checkpoint. Intervention analysis proceeds only
after this reconstructed reference condition reproduces the corresponding
stored inference outputs and metrics.

Let $q_{t-1}$ denote the recurrence-visible state entering recurrent step
$t$, $x_t$ the current input, $z_t$ the update gate, $r_t$ the reset
gate, $\tilde{h}_t$ the candidate state, and $h_t$ the raw hidden state
computed before storage. The time indexing follows the storage sequence
$q_{t-1}\rightarrow h_t\rightarrow q_t$: the recurrence-visible state stored
after step $t-1$ is read at step $t$, the GRU computes the raw state $h_t$,
and the write-back operator produces $q_t$, which is then supplied to
step $t+1$. The implemented GRU recurrence is
\begin{align}
z_t
&=
\sigma\left(
x_tW_z+q_{t-1}U_z+b_z
\right), \\
r_t
&=
\sigma\left(
x_tW_r+q_{t-1}U_r+b_r
\right), \\
\tilde{h}_t
&=
\phi_a\left(
x_tW_h+r_t\odot(q_{t-1}U_h)+b_h
\right), \\
h_t
&=
z_t\odot q_{t-1}
+
(1-z_t)\odot\tilde{h}_t ,
\label{eq:supp_gru_recurrence}
\end{align}
where $\odot$ denotes element-wise multiplication.

Continuous state propagation returns the computed state directly,
$q_t=h_t$. Deterministic $B$-bit state write-back instead stores
\begin{equation}
q_t
=
Q_B(h_t),
\label{eq:supp_deterministic_writeback}
\end{equation}
where $Q_B$ denotes nearest-level quantization on a state grid with spacing
$\Delta_B$.

The state change is
\begin{equation}
\delta_t
=
h_t-q_{t-1}
=
(1-z_t)\odot
\left(
\tilde{h}_t-q_{t-1}
\right).
\label{eq:supp_requested_state_change}
\end{equation}
For hidden unit $j$, the recurrent write margin is therefore
\begin{equation}
M_{t,j}
=
\frac{2|\delta_{t,j}|}{\Delta_B}
=
\frac{
2(1-z_{t,j})
|\tilde{h}_{t,j}-q_{t-1,j}|
}{
\Delta_B
}.
\label{eq:supp_write_margin}
\end{equation}
At an interior grid level, $M_{t,j}<1$ lies within the deterministic
half-step write boundary and retains the same recurrence-visible state.
Exact half-step ties are rounded to the nearest even quantization index. Values outside the
representable state range are clipped to the nearest state-grid rail.

For error-feedback write-back,
\begin{align}
q_t
&=
Q_B(h_t+e_{t-1}), \\
e_t
&=
\operatorname{clip}
\left(
h_t+e_{t-1}-q_t,
-\Delta_B,
+\Delta_B
\right).
\end{align}
The error-feedback residual is an auxiliary write-back state rather than a
trained network variable.

For $k$-bit quantized residual memory, let $\rho_t$ denote the auxiliary
stored residual. The compensated write-back is
\begin{align}
q_t
&=
Q_B\left(h_t+\rho_{t-1}\right), \\
u_t
&=
h_t+\rho_{t-1}-q_t, \\
\rho_t
&=
Q_{\rho,k}(u_t),
\end{align}
where $Q_{\rho,k}$ contains $2^k$ uniformly spaced levels over
$[-\Delta_B/2,\Delta_B/2)$ with spacing
\begin{equation}
\Delta_{\rho,k}
=
\frac{\Delta_B}{2^k}.
\end{equation}
The evaluated auxiliary residual widths are $k=2,3,4$. The
floating-point residual reference retains the residual over
$[-\Delta_B/2,+\Delta_B/2]$. Error feedback uses the separate auxiliary
state $e_t$ defined above.

Stochastic rounding selects adjacent representable state levels according to
the relative position of the raw state between them. Five independent
stochastic inference realizations are evaluated for each reported stochastic
condition.

For $k$-bit direction memory, let $\kappa_t$ denote the auxiliary direction counter and define the trigger magnitude

\begin{equation}
T_k
=
2^{k-1}.
\label{eq:counter_trigger}
\end{equation}
The stored counter takes values in
\begin{equation}
\mathcal{K}_k
=
\left\{
-(T_k-1),\ldots,-1,0,1,\ldots,T_k-1
\right\}.
\label{eq:counter_states}
\end{equation}
Because
$|\mathcal{K}_k|=2T_k-1=2^k-1$, these active counter states occupy
$2^k-1$ of the $2^k$ binary codes available in a $k$-bit register. The two
values $\pm T_k$ are trigger conditions and are not stored counter states.

At recurrent step $t$, the proposed state change is
\begin{equation}
\delta_t
=
h_t-q_{t-1}.
\end{equation}

If
\begin{equation}
|\delta_t|
\geq
\frac{\Delta_B}{2},
\end{equation}
the proposed state change reaches the deterministic write boundary. The
operator therefore applies ordinary nearest-level write-back,
$q_t=Q_B(h_t)$, and resets the direction-memory counter.

For a sub-threshold proposed state change satisfying
\begin{equation}
\frac{\Delta_B}{8}
<
|\delta_t|
<
\frac{\Delta_B}{2},
\label{eq:counter_vote_region}
\end{equation}
the direction-memory counter receives the signed vote
\begin{equation}
v_t
=
\operatorname{sign}(\delta_t),
\end{equation}
and the candidate counter value is
\begin{equation}
\kappa_t^{\prime}
=
\kappa_{t-1}+v_t.
\end{equation}

If $\kappa_t^{\prime}\geq T_k$, the recurrence-visible state advances by one
positive grid step and the counter is reset. If
$\kappa_t^{\prime}\leq -T_k$, the recurrence-visible state advances by one
negative grid step and the counter is reset. State transitions are clipped
at the representable grid rails. If neither trigger is reached, the stored
state remains $q_t=q_{t-1}$ and the updated counter is retained,
$\kappa_t=\kappa_t^{\prime}$. proposed state changes satisfying
$|\delta_t|\leq\Delta_B/8$ retain the existing counter without adding a
vote.

The evaluated direction-memory widths are $k=2,3,4$.

Encoder and decoder auxiliary states remain independent. For quantized
residual and full-half-step residual conditions, auxiliary residual memory
is reset before decoder handoff, after which the residual generated by
handoff quantization initializes the decoder residual. Direction-memory counter state is reset at the encoder-to-decoder boundary.

\subsubsection*{Representative-trajectory selection}
\label{sec:supp_trajectory_selection}

For the representative trajectory in Fig.~\ref{fig:write-back_mechanism}a,
selection was rule based and did not use visual inspection. Among held-out
sequences containing at least one recurrence-visible direction-memory trigger, we
selected the sequence whose deterministic 8-bit-to-4-bit decoder state-change
fraction was closest to the median over the complete held-out test set, with
ties resolved by the lowest test-partition position. Within that sequence,
we selected the decoder unit with the greatest number of recurrence-visible
direction-memory triggers, with ties resolved first by the largest deterministic
active-vote count and then by the lowest unit index. This procedure selected
test-partition position 9 and decoder unit 21.

\ subsubsection* {Proposed state change persistence and direction-memory trigger analysis}
\label{sec:supp_sign_persistence}

To characterize temporal persistence, we analyze decoder proposed state
changes at the P3 checkpoint, the 4-bit-state reference GRU, and the
8-bit-state reference GRU under post-training 4-bit write-back. Direction
memory is evaluated for the reference-model analyses using the fixed voting
region $\Delta_B/8<|\delta_t|<\Delta_B/2$. Trained network parameters remain fixed throughout the post-training interventions.

For each decoder write opportunity, the proposed state change is
\begin{equation}
\delta_t = h_t-q_{t-1}.
\end{equation}
A proposed state change contributes a direction-memory vote when
\begin{equation}
\frac{\Delta_B}{8}
<
|\delta_t|
<
\frac{\Delta_B}{2},
\end{equation}
with vote sign
\begin{equation}
v_t=\operatorname{sign}(\delta_t).
\end{equation}
Only decoder steps that can update recurrence-visible state are included.
The terminal decoder output is excluded because its raw hidden state is used
for readout but is not written back for a subsequent recurrent step.

One-step sign persistence is calculated from consecutive eligible vote
events. Let $n_{++}$, $n_{+-}$, $n_{-+}$, and $n_{--}$ denote the numbers of
positive-to-positive, positive-to-negative, negative-to-positive, and
negative-to-negative adjacent vote transitions, respectively. The same-sign
fraction is
\begin{equation}
P_{\mathrm{same}}
=
\frac{n_{++}+n_{--}}
{n_{++}+n_{+-}+n_{-+}+n_{--}}.
\end{equation}

Same-sign run length is defined as the number of consecutive decoder write opportunities that cast eligible votes with the same sign. A run terminates
when the vote sign reverses, when no eligible vote is cast, when an ordinary
above-half-step state write occurs, when direction-memory triggers a state transition, or
when the sequence ends. This run definition measures uninterrupted temporal
persistence and is distinct from the direction-memory counter state itself.

Sign-transition statistics are accumulated independently for each held-out
sequence and summarized over the complete 160,000-sample test partition. Run-length distributions
and direction-memory trigger probabilities are calculated from the corresponding pooled
decoder events. All recurrent trajectories are reconstructed from frozen
model parameters without retraining.

For state-grid spacing $\Delta_B$, the recurrence-visible write is
\begin{equation}
W_t =
\frac{|q_t-q_{t-1}|}{\Delta_B}.
\end{equation}
The element-wise zero-write probability,
\begin{equation}
P(W_t=0),
\end{equation}
measures how frequently an individual hidden unit remains on the same stored
state level. Vector-level write activity is
\begin{equation}
N_{\mathrm{write}}(t)
=
\sum_{j=1}^{32}
\mathbf{1}
\left[
q_{t,j}\neq q_{t-1,j}
\right].
\end{equation}
We report $P(N_{\mathrm{write}}=0)$ and the mean number of changing hidden
units per recurrent step.

For discrete stateful operators, the conditional sub-threshold write
fraction is
\begin{equation}
P
\left(
q_t\neq q_{t-1}
\,\middle|\,
M_t<1
\right).
\end{equation}
This statistic measures the fraction of proposed state changes lying within
the deterministic half-step deadband that nevertheless become
recurrence-visible transitions under the corresponding stateful write-back
operator.

For each hidden unit, state occupancy is summarized by the number of
discrete levels visited and by
\begin{equation}
N_{\mathrm{eff}}=\exp(H),
\qquad
H=-\sum_{\ell} p_{\ell}\log p_{\ell},
\end{equation}
where $p_{\ell}$ is the empirical occupancy probability of state level $\ell$.
$N_{\mathrm{eff}}$ summarizes concentration of the marginal occupancy
distribution and is not interpreted as an information capacity or effective
network bit width.

The encoder-to-decoder handoff error is reported as the mean absolute
difference between the encoder final state and the decoder initial
recurrence-visible state after application of the corresponding handoff
write-back. Rail occupancy is reported separately to distinguish state-grid
persistence from clipping.

\subsubsection*{LSTM cross-architecture replication and state-type dissection}
\label{sec:supp_lstm}

The cross-architecture analysis uses an independently trained 32-unit LSTM
checkpoint with 8-bit kernels, recurrent kernels, biases, activations, and
native recurrent states. Let $x_t^{\mathrm{raw}}$ denote a raw LSTM
recurrent state proposal for $x\in\{c,h\}$ and let $q_t^{(x)}$ denote the
corresponding recurrence-visible stored state. Deterministic state-specific
write-back is
\begin{equation}
q_t^{(x)}
=
Q_{B_x}
\left(
x_t^{\mathrm{raw}}
\right),
\qquad
x\in\{c,h\}.
\label{eq:lstm_state_write-back}
\end{equation}
The native condition uses $B_c=B_h=8$. Whole-state deterministic 4-bit uses
$B_c=B_h=4$. The $c$-only intervention uses $B_c=4$ and $B_h=8$, while the
$h$-only intervention uses $B_c=8$ and $B_h=4$. Identity propagation returns
the raw recurrent state directly without discrete write-back.

For targeted error-feedback conditions, the same error-feedback operator
used in the GRU analysis is applied only to the selected 4-bit LSTM state
variable. For $x\in\{c,h\}$,
\begin{align}
q_t^{(x)}
&=
Q_4
\left(
x_t^{\mathrm{raw}}+e_{t-1}^{(x)}
\right), \\
e_t^{(x)}
&=
\operatorname{clip}
\left(
x_t^{\mathrm{raw}}+e_{t-1}^{(x)}-q_t^{(x)},
-\Delta_4,
+\Delta_4
\right).
\end{align}
The non-targeted recurrent state retains deterministic 8-bit write-back. The
specified write-back condition is applied in both encoder and decoder
recurrence. No LSTM checkpoint is retrained or fine-tuned for these
interventions.

The encoder contains 135 live recurrent-state write opportunities per
sequence and the decoder contains 134. As in the GRU analysis, the terminal
decoder output is excluded from write statistics because it is not returned
to a subsequent recurrent step. Deadband fraction, state-change fraction,
conditional sub-threshold write fraction, same-sign fraction, and completed same-sign run lengths are accumulated separately
for the LSTM cell and hidden states using their corresponding grids.

The reconstructed native LSTM was first compared with native inference on
2,048 samples and then on all 160,000 held-out test samples. Both
comparisons gave maximum absolute difference 0.0 and mean absolute
difference 0.0 at a predefined tolerance of $5\times10^{-5}$. 

\subsubsection*{Matched recurrent-memory training campaign}
\label{sec:supp_trained_campaign}

The trained campaign compares four recurrent-memory allocations, with each
network trained using its assigned recurrent-memory interface throughout
optimization: a 4-bit recurrence-visible state with no auxiliary memory
(4 stored bits per unit), a 6-bit recurrence-visible state with no auxiliary
memory (6 stored bits per unit), a 4-bit recurrence-visible state with 2-bit
residual memory (6 stored bits per unit), and a 4-bit recurrence-visible
state with 2-bit direction memory (6 stored bits per unit). Direction memory
uses a 2-bit auxiliary direction counter with trigger magnitude $T_2=2$. Input kernels,
recurrent kernels, biases, and candidate activation remain 4-bit in all four
conditions. Thus, the 6-bit-state condition changes only the precision of the
recurrence-visible state, while the residual-memory and direction-memory
conditions retain a 4-bit recurrence-visible state and allocate the
additional two stored bits to auxiliary memory.

The campaign specification was fixed before training. All conditions share
the same teacher, data partition, knowledge-distillation objective with
$\alpha=0.6$, optimizer, learning-rate schedule, batch configuration, and
stopping rule. Three matched training runs are evaluated for each condition.
Within each run, the initial trainable parameters are identical across the
four recurrent-memory conditions.

For the residual-memory and direction-memory conditions, the forward recurrence-visible
state is hard discrete during training. The visible state uses an identity
straight-through surrogate. Auxiliary direction and residual memory are stop-gradient recurrent states: the direction-memory condition has no gradient through
vote accumulation or trigger decisions, while the residual condition
retains a local amplitude-dependent surrogate through the compensated
visible-state write. Test evaluation uses the common 160,000-sample held-out
partition with the lifetime extraction procedure of Supplementary
Methods~\ref{sec:supp_lifetime}. Because the campaign uses its own fixed
training budget, its runs are compared only within the campaign and not
against the reference models of Supplementary
Table~\ref{tab:temporal_memory_sweep}.

\subsection{Lifetime estimation and reconstruction fidelity}
\label{sec:supp_lifetime}
The model predicts three temporal output parameters; the present analysis focuses on the two parameters carrying fluorescence lifetime decay information. The physical time
vector is
\begin{equation}
t_n = n\Delta t,
\qquad
\Delta t=0.09~\mathrm{ns},
\qquad
n=0,\ldots,134.
\end{equation}

For the two predicted lifetime parameters $c\in\{1,2\}$, the lifetime estimate is obtained by trapezoidal integration normalized to the first-gate amplitude,
\begin{equation}
\hat{\tau}_{c,i}
=
\frac{
\operatorname{trapz}\left[s_{c,i}(t),t\right]
}{
s_{c,i}(0)
},
\label{eq:lifetime}
\end{equation}
when $s_{c,i}(0)>10^{-6}$; otherwise the implementation returns zero. The sample-level estimated parameter vector is therefore
\begin{equation}
\hat{\boldsymbol{\theta}}_i
=
\left(
\hat{\tau}_{1,i},
\hat{\tau}_{2,i}
\right).
\end{equation}
Ground-truth $\tau_{1,i}$ and $\tau_{2,i}$ are taken directly from the stored
scalar label array.

For lifetime component $c\in\{1,2\}$,
\begin{equation}
\mathrm{RMSE}_{\tau_c}
=
\sqrt{
\frac{1}{N}
\sum_{i=1}^{N}
\left(
\hat{\tau}_{c,i}
-
\tau_{c,i}
\right)^2
},
\end{equation}
and $r_{\tau_c}$ denotes the Pearson correlation between the estimated and
ground-truth lifetime values across samples.

Predicted-versus-ground-truth distributions are reported alongside aggregate
metrics because quantization can produce clustering at discrete prediction
levels, dynamic-range compression, or reduced input dependence that a single
error statistic may not reveal.

\begin{table*}[ht!]
\centering
\caption{Aggregate recurrent-state write-back statistics for principal native
and intervention conditions. $P(W_0)$ is the element-wise zero-write
probability, $P(N_0)$ is the probability that no hidden unit changes state
level during a recurrent step, and mean $N$ is the average number of
changing hidden units per step.}
\label{tab:write-back_summary}
\scriptsize
\setlength{\tabcolsep}{3.2pt}
\renewcommand{\arraystretch}{1.08}

\resizebox{\textwidth}{!}{%
\begin{tabular}{lccccccc}
\toprule
Condition
& \multicolumn{3}{c}{Encoder}
& \multicolumn{3}{c}{Decoder}
& Handoff MAE \\
\cmidrule(lr){2-4}
\cmidrule(lr){5-7}

& $P(W_0)$
& $P(N_0)$
& Mean $N$
& $P(W_0)$
& $P(N_0)$
& Mean $N$
& \\
\midrule

P2F native
& 0.9536 & 0.4746 & 1.4859
& 0.9975 & 0.9725 & 0.0813
& 0.015483 \\

P2F error feedback
& 0.8945 & 0.1675 & 3.3749
& 0.9489 & 0.6945 & 1.6351
& 0.000837 \\

P2F stochastic
& 0.9043 & 0.2190 & 3.0618
& 0.9538 & 0.6860 & 1.4789
& 0.001335 \\

P3 native
& 0.8858 & 0.3107 & 3.6556
& 0.9325 & 0.5807 & 2.1593
& 0.008159 \\

P3 error feedback
& 0.8643 & 0.1342 & 4.3439
& 0.7918 & 0.0020 & 6.6620
& 0.001574 \\

P3 stochastic
& 0.8664 & 0.1443 & 4.2750
& 0.8801 & 0.1297 & 3.8361
& 0.002840 \\

Native 4-bit
& 0.9241 & 0.4057 & 2.4296
& 0.8753 & 0.1918 & 3.9893
& 0.008928 \\

Native 8-bit
& 0.3681 & 0.0003 & 20.2208
& 0.6089 & 0.2195 & 12.5148
& 0.001925 \\

8-bit-state reference, post-training 4-bit
& 0.9654 & 0.6346 & 1.1070
& 0.9940 & 0.9794 & 0.1915
& 0.028255 \\

\bottomrule
\end{tabular}%
}
\end{table*}

\begin{table*}[ht!]
\centering
\caption{Median per-unit state occupancy and write-back statistics for
selected conditions. Occupied levels count every state level visited by a
hidden unit; $N_{\mathrm{eff}}$ summarizes occupancy entropy.}
\label{tab:per_unit_summary}
\scriptsize
\setlength{\tabcolsep}{4.0pt}
\renewcommand{\arraystretch}{1.08}

\resizebox{0.95\textwidth}{!}{%
\begin{tabular}{llcccc}
\toprule
Condition
& Region
& $N_{\mathrm{eff}}$
& Occupied levels
& $P(W_0)$
& Rail fraction \\
\midrule

P2F native
& Decoder & 1.386 & 4.0 & 0.9994 & 0.0000 \\
P2F native
& Encoder & 1.920 & 5.0 & 0.9794 & 0.0000 \\

Native 4-bit
& Decoder & 3.427 & 12.0 & 0.9114 & 0.0093 \\
Native 4-bit
& Encoder & 3.185 & 8.0 & 0.9473 & 0.1515 \\

Native 4-bit with 8-bit state
& Decoder & 6.673 & 176.5 & 0.7499 & 0.0000 \\
Native 4-bit with 8-bit state
& Encoder & 8.468 & 112.5 & 0.8414 & 0.0000 \\

Native 8-bit
& Decoder & 39.734 & 154.0 & 0.5854 & 0.0000 \\
Native 8-bit
& Encoder & 112.021 & 149.5 & 0.3231 & 0.0000 \\

8-bit-state reference, post-training 4-bit
& Decoder & 1.109 & 5.0 & 0.9937 & 0.0000 \\
8-bit-state reference, post-training 4-bit
& Encoder & 2.479 & 4.0 & 0.9828 & 0.0000 \\

\bottomrule
\end{tabular}%
}
\end{table*}

\begin{table*}[ht!]
\centering
\caption{\textbf{Selected post-training temporal-memory interventions.}
All conditions use the common 160,000-sample held-out test partition.
Stochastic-rounding RMSE values are mean $\pm$ standard deviation across
five inference realizations; other rows are deterministic evaluations.
$P_{\mathrm{sub}\rightarrow\mathrm{write}}$ is the decoder fraction of
proposed state changes with $M_t<1$ that nevertheless produce a
recurrence-visible state transition. It is not defined for identity
propagation because no discrete state write-back operator is active.}
\label{tab:temporal_memory_sweep}
\scriptsize
\setlength{\tabcolsep}{3.0pt}
\renewcommand{\arraystretch}{1.08}

\resizebox{\textwidth}{!}{%
\begin{tabular}{llccccc}
\toprule
Checkpoint
& State write-back
& Stored state + auxiliary bits
& Sequence MAE
& $\tau_1$ RMSE
& $\tau_2$ RMSE
& $P_{\mathrm{sub}\rightarrow\mathrm{write}}$ \\
\midrule

P2F
& Identity
& continuous
& 0.023940
& 0.360333
& 0.351951
& -- \\

P2F
& Deterministic 4-bit
& 4
& 0.091492
& 25.368287
& 106.589924
& 0.000000 \\

P2F
& Error feedback
& 4 + float residual
& 0.026176
& 0.358926
& 0.369030
& 0.032141 \\

P2F
& Stochastic rounding
& 4
& 0.030626
& $0.3784\pm0.0003$
& $0.4250\pm0.0002$
& -- \\

P2F
& 2-bit residual memory
& 4 + 2
& 0.025052
& 0.336577
& 0.398735
& 0.040461 \\

P2F
& 3-bit direction memory
& 4 + 3
& 0.025630
& 0.335884
& 0.336644
& 0.020589 \\

\midrule

P3
& Identity
& continuous
& 0.061813
& 0.394100
& 0.444563
& -- \\

P3
& Deterministic 4-bit
& 4
& 0.042935
& 0.479972
& 0.553816
& 0.000000 \\

P3
& Error feedback
& 4 + float residual
& 0.058720
& 0.379709
& 0.489615
& 0.110312 \\

\midrule

Native 4-bit
& Deterministic 4-bit
& 4
& 0.028412
& 0.347913
& 0.401108
& 0.000000 \\

Native 4-bit
& Error feedback
& 4 + float residual
& 0.038218
& 0.423251
& 0.546330
& 0.029586 \\

Native 4-bit
& 4-bit residual memory
& 4 + 4
& 0.037620
& 0.420722
& 0.538478
& 0.032247 \\

Native 4-bit
& 4-bit direction memory
& 4 + 4
& 0.036431
& 0.359385
& 0.463926
& 0.003950 \\

\midrule

Native 8-bit
& Deterministic 8-bit
& 8
& 0.014685
& 0.202834
& 0.216423
& 0.000000 \\

Native 8-bit
& Post-training deterministic 4-bit
& 4
& 0.102485
& 1.887303
& 2.603046
& 0.000000 \\

Native 8-bit
& Post-training 4-bit + error feedback
& 4 + float residual
& 0.046218
& 0.340835
& 0.459084
& 0.324496 \\

Native 8-bit
& Post-training 4-bit + 2-bit residual memory
& 4 + 2
& 0.091862
& 0.511585
& 1.514574
& 0.149631 \\

Native 8-bit
& Post-training 4-bit + 4-bit residual memory
& 4 + 4
& 0.046230
& 0.340557
& 0.453148
& 0.315198 \\

Native 8-bit
& Post-training 4-bit + 4-bit direction memory
& 4 + 4
& 0.052654
& 0.485008
& 0.575160
& 0.031249 \\

\bottomrule
\end{tabular}%
}
\end{table*}

\begin{table*}[ht!]
\centering
\caption{\textbf{Decoder recurrent write-margin statistics for selected
conditions.} Deadband fraction is $P(M_t<1)$. Identity margins are
counterfactual measurements relative to the indicated state-grid spacing.
State-change fraction is the fraction of decoder state elements whose
recurrence-visible value changes between consecutive recurrent steps.}
\label{tab:margin_summary}
\scriptsize
\setlength{\tabcolsep}{4.0pt}
\renewcommand{\arraystretch}{1.08}

\resizebox{0.95\textwidth}{!}{%
\begin{tabular}{lcccccc}
\toprule
Condition
& State bits
& Deadband fraction
& Counterfactual
& State-change fraction
& Margin $p_{90}$
& Margin $p_{99}$ \\
\midrule

P2E identity
& 4
& 0.974994
& Yes
& 0.358225
& 0.215086
& 1.972369 \\

P2E deterministic 4-bit
& 4
& 0.996892
& No
& 0.003044
& 0.842488
& 0.974532 \\

P2F identity
& 4
& 0.982196
& Yes
& 0.297849
& 0.242823
& 1.522765 \\

P2F deterministic 4-bit
& 4
& 0.995911
& No
& 0.002542
& 0.927245
& 0.982090 \\

P3 identity
& 4
& 0.939679
& Yes
& 0.400494
& 0.646396
& 2.251092 \\

P3 deterministic 4-bit
& 4
& 0.930550
& No
& 0.067477
& 0.941925
& 2.181308 \\

Native 4-bit deterministic
& 4
& 0.865438
& No
& 0.124666
& 1.090997
& 2.247787 \\

Native 8-bit deterministic
& 8
& 0.608850
& No
& 0.391089
& 2.946978
& 21.130625 \\

8-bit-state reference, post-training 4-bit
& 4
& 0.994017
& No
& 0.005983
& 0.763094
& 0.978958 \\

\bottomrule
\end{tabular}%
}
\end{table*}

\begin{table*}[ht!]
\centering
\caption{\textbf{Decoder proposed state change persistence summary.}
Statistics are accumulated over the complete 160,000-sample held-out test
partition following Supplementary
Methods~\ref{sec:supp_sign_persistence}. Deadband fraction is $P(M_t<1)$,
state-change fraction is the fraction of decoder state elements whose
recurrence-visible value changes between consecutive recurrent steps,
$P_{\mathrm{sub}\rightarrow\mathrm{write}}$ is the conditional sub-threshold
write fraction, and the same-sign fraction is $P_{\mathrm{same}}$ over
adjacent eligible vote pairs. Deterministic rows report proposed state change statistics on the
deterministic trajectory, where no auxiliary direction counter operates.
Direction-memory rows use a 4-bit auxiliary counter and the fixed voting
region $\Delta_B/8<|\delta_t|<\Delta_B/2$.
}
\label{tab:sign_persistence}
\scriptsize
\setlength{\tabcolsep}{4.0pt}
\renewcommand{\arraystretch}{1.08}

\resizebox{0.95\textwidth}{!}{%
\begin{tabular}{llcccc}
\toprule
Condition
& Operator
& Deadband fraction
& State-change fraction
& $P_{\mathrm{sub}\rightarrow\mathrm{write}}$
& Same-sign fraction \\
\midrule

P3
& Deterministic 4-bit
& 0.930550 & 0.067477 & 0.000000 & 0.99969 \\

Native 4-bit
& Deterministic 4-bit
& 0.865438 & 0.124666 & 0.000000 & 0.99943 \\

Native 4-bit
& 4-bit direction memory
& 0.904323 & 0.096554 & 0.003950 & 0.99894 \\

8-bit-state reference, post-training 4-bit
& Deterministic 4-bit
& 0.994017 & 0.005983 & 0.000000 & 0.99976 \\

8-bit-state reference, post-training 4-bit
& 4-bit direction memory
& 0.983090 & 0.047625 & 0.031249 & 0.99917 \\

\bottomrule
\end{tabular}%
}
\end{table*}
\begin{table*}[ht!]
\centering
\caption{\textbf{Matched recurrent-memory training campaign.}
Lifetime RMSE is reported in nanoseconds on the complete 160,000-sample
held-out test partition for three matched training runs per recurrent-memory
condition. Within each run, all four conditions start from identical
trainable parameters and share the training specification described in
Supplementary Methods~\ref{sec:supp_trained_campaign}. Mean $\pm$ SD
summarizes variation across the three matched training runs.}
\label{tab:trained_campaign}
\scriptsize
\setlength{\tabcolsep}{3.0pt}
\renewcommand{\arraystretch}{1.08}

\resizebox{\textwidth}{!}{%
\begin{tabular}{lcccccccc}
\toprule
Condition
& $\tau_1$ run 1
& $\tau_1$ run 2
& $\tau_1$ run 3
& $\tau_1$ mean $\pm$ SD
& $\tau_2$ run 1
& $\tau_2$ run 2
& $\tau_2$ run 3
& $\tau_2$ mean $\pm$ SD \\
\midrule

4-bit state
& 0.374929
& 0.432975
& 0.427440
& $0.411782 \pm 0.032035$
& 0.445717
& 0.532873
& 0.460267
& $0.479619 \pm 0.046690$ \\

6-bit state
& 0.298677
& 0.566732
& 0.577438
& $0.480949 \pm 0.157943$
& 0.278720
& 0.526481
& 0.339875
& $0.381692 \pm 0.129065$ \\

4-bit state + 2-bit residual memory
& 0.435202
& 0.429969
& 0.263304
& $0.376158 \pm 0.097770$
& 0.342398
& 0.474813
& 0.276072
& $0.364427 \pm 0.101186$ \\

4-bit state + 2-bit direction memory
& 0.288118
& 0.280328
& 0.328112
& $0.298853 \pm 0.025637$
& 0.343285
& 0.581813
& 0.374296
& $0.433131 \pm 0.129692$ \\

\bottomrule
\end{tabular}%
}
\end{table*}
\begin{table*}[ht!]
\centering
\caption{\textbf{Paired differences across matched recurrent-memory
training runs.}
Differences are calculated as the lifetime RMSE of the first-listed
condition minus that of the second-listed condition for the matched
training runs defined in Supplementary
Methods~\ref{sec:supp_trained_campaign}. Negative values therefore indicate lower
RMSE for the first-listed condition. The main text emphasizes comparisons
whose direction is consistent across all three matched runs.}
\label{tab:trained_campaign_paired}
\scriptsize
\setlength{\tabcolsep}{3.0pt}
\renewcommand{\arraystretch}{1.08}

\resizebox{\textwidth}{!}{%
\begin{tabular}{lcccccccc}
\toprule
Comparison
& $\Delta\mathrm{RMSE}_{\tau_1}$ run 1
& $\Delta\mathrm{RMSE}_{\tau_1}$ run 2
& $\Delta\mathrm{RMSE}_{\tau_1}$ run 3
& $\Delta\mathrm{RMSE}_{\tau_1}$ mean
& $\Delta\mathrm{RMSE}_{\tau_2}$ run 1
& $\Delta\mathrm{RMSE}_{\tau_2}$ run 2
& $\Delta\mathrm{RMSE}_{\tau_2}$ run 3
& $\Delta\mathrm{RMSE}_{\tau_2}$ mean \\
\midrule

6-bit state $-$ 4-bit state
& $-0.076253$ & $+0.133756$ & $+0.149998$ & $+0.069167$
& $-0.166997$ & $-0.006392$ & $-0.120392$ & $-0.097927$ \\

4-bit state + 2-bit residual memory $-$ 4-bit state
& $+0.060272$ & $-0.003006$ & $-0.164136$ & $-0.035623$
& $-0.103319$ & $-0.058060$ & $-0.184196$ & $-0.115192$ \\

4-bit state + 2-bit direction memory $-$ 4-bit state
& $-0.086811$ & $-0.152647$ & $-0.099328$ & $-0.112929$
& $-0.102431$ & $+0.048940$ & $-0.085971$ & $-0.046487$ \\

4-bit state + 2-bit residual memory $-$ 6-bit state
& $+0.136525$ & $-0.136762$ & $-0.314134$ & $-0.104790$
& $+0.063678$ & $-0.051668$ & $-0.063804$ & $-0.017265$ \\

4-bit state + 2-bit direction memory $-$ 6-bit state
& $-0.010559$ & $-0.286403$ & $-0.249327$ & $-0.182096$
& $+0.064565$ & $+0.055332$ & $+0.034421$ & $+0.051439$ \\

4-bit state + 2-bit direction memory $-$
4-bit state + 2-bit residual memory
& $-0.147084$ & $-0.149641$ & $+0.064807$ & $-0.077306$
& $+0.000887$ & $+0.107000$ & $+0.098224$ & $+0.068704$ \\

\bottomrule
\end{tabular}%
}
\end{table*}

\subsection{Paired storage and lifetime analyses}
\label{sec:supp_paired_statistics}

For the native-4-bit matched-storage analysis, deterministic $(4+k)$-bit state write-back is compared with a 4-bit recurrence-visible state plus a $k$-bit quantized residual memory or $k$-bit direction memory, with $k\in\{2,3,4\}$. The
resulting conditions store 6, 7, or 8 recurrent-memory bits per hidden unit.
This comparison concerns stored recurrent memory only and does not imply
equality of arithmetic, logic, routing, energy, or total model memory.

Paired uncertainty is calculated at the held-out sequence level. Let
$m_A(\mathcal{S})$ and $m_B(\mathcal{S})$ denote the same metric evaluated
for two frozen write-back conditions on the same resampled set of held-out
sequences $\mathcal{S}$. For bootstrap replicate $b$,
\begin{equation}
D^{(b)}
=
m_B(\mathcal{S}^{(b)})
-
m_A(\mathcal{S}^{(b)}).
\end{equation}
Two thousand paired bootstrap replicates are used. The reported 95\%
confidence interval is the interval between the 2.5th and 97.5th percentiles
of the paired-difference distribution.

For lifetime-conditioned analysis, the held-out samples are sorted
separately by ground-truth $\tau_1$ and ground-truth $\tau_2$ and divided
into ten equal-count bins of 16,000 samples. Identity and deterministic
predictions from the same checkpoint are compared using paired
sequence-level bootstrap resampling within each bin. The bin definitions
therefore depend only on ground-truth lifetime and not on prediction error.

\subsection{Software environment and reproducibility}
\label{sec:supp_reproducibility}

Training and inference use TensorFlow with QKeras. The primary QMem
configuration fixes the reported 4-bit precision and phase schedule.
Native tensor and metric fidelity checks are required before each
post-training intervention analysis is accepted. The QMem hardening
sequence represents one training trajectory, whereas the native 4-bit and
native 8-bit GRU models are independently trained reference solutions.

\subsection*{Data and code availability}

No human or live-animal dataset
is analyzed in the present work. The complete training and analysis code and
trained checkpoints for all reported results will be released publicly upon
acceptance, together with the dataset; links are withheld here to preserve
anonymity under double-blind review.

\subsection*{Ethics statement}

Not applicable. The present study does not analyze human participants, live
animal subjects, or biological-specimen experimental datasets.

\paragraph*{LLM usage disclosure}
Large language models were used to assist with editing the manuscript
prose and with layout of conceptual figure components (the schematic
panels of Figs.~\ref{fig:overview} and \ref{fig:supp_qmem}). No results,
data, analyses, metrics, or models were generated with these tools.

\clearpage
\bibliographystyle{unsrtnat}
\bibliography{cas-refs}

\end{document}